\documentclass[lettersize,journal]{IEEEtran}
\usepackage{amsmath,amssymb,amsfonts}
\usepackage{amsthm}
\usepackage{algorithmic}
\usepackage{algorithm}
\usepackage{array}
\usepackage[caption=false,font=normalsize,labelfont=sf,textfont=sf]{subfig}
\usepackage{textcomp}
\usepackage{stfloats}

\usepackage{url}
\usepackage{verbatim}
\usepackage{amssymb}
\usepackage{pifont}
\usepackage{graphicx}

\usepackage{cite}
\newtheorem{theorem}{Theorem}

\usepackage{graphicx}  

\usepackage{multirow}
\usepackage{booktabs}   
\def\BibTeX{{\rm B\kern-.05em{\sc i\kern-.025em b}\kern-.08em
    T\kern-.1667em\lower.7ex\hbox{E}\kern-.125emX}}
\usepackage{balance}

\usepackage{hyperref}
\hypersetup{hypertex=true,
            colorlinks=true,
            linkcolor=blue,
            anchorcolor=blue,
            citecolor=blue}

\begin{document}

\title{Dataset Distillation-based Hybrid Federated Learning on Non-IID Data}

\author{Xiufang Shi,~\IEEEmembership{Member,~IEEE,}
Wei Zhang, Yuheng Li,  Mincheng Wu,~\IEEEmembership{Member,~IEEE,} Zhenyu Wen,~\IEEEmembership{Senior Member,~IEEE,} Shibo He,~\IEEEmembership{Senior Member,~IEEE},
Tejal Shah, Rajiv Ranjan,~\IEEEmembership{Fellow,~IEEE}

\thanks{Xiufang Shi, Wei Zhang, Yuheng Li, Mincheng Wu and Zhenyu Wen are with the College of Information Engineering, Zhejiang University of Technology, Hangzhou 310023, China, and also with the Zhejiang Key Laboratory of Intelligent Perception and Control for Complex Systems, Hangzhou, China. (e-mail: xiufangshi@zjut.edu.cn; 221122030290@zjut.edu.cn; 221124030326@zjut.edu.cn; minchengwu@zjut.edu.cn; wenluke427@gmail.com). 
Shibo He is with the College of Control Science and Engineering, Zhejiang University, Hangzhou 310027, China (e-mail: s18he@zju.edu.cn).
Tejal Shah and Rajiv Ranjan are with Computing Science and Internet of Things, Newcastle University, NE1 7RU Newcastle, UK (e-mail: tejal.shah@newcastle.ac.uk; raj.ranjan@newcastle.ac.uk)
(Corresponding author: Mincheng Wu.)
}
}


\markboth{
}%
{Shell \MakeLowercase{\textit{et al.}}: A Sample Article Using IEEEtran.cls for IEEE Journals}

\maketitle


\begin{abstract}
In federated learning, the heterogeneity of client data has a great impact on the performance of model training. 
Many heterogeneity issues in this process are raised by non-independently and identically distributed (non-IID) data. To address the issue of label distribution skew, we propose a hybrid federated learning framework called HFLDD, which integrates dataset distillation to generate approximately independent and equally distributed (IID) data, thereby improving the performance of model training. In particular, 
we partition the clients into heterogeneous clusters, where the data labels among different clients within a cluster are unbalanced while the data labels among different clusters are balanced. 
The cluster heads collect distilled data from the corresponding cluster members, and conduct model training in collaboration with the server. 
This training process is like traditional federated learning on IID data, and hence effectively alleviates the impact of non-IID data on model training. We perform a comprehensive analysis of the convergence behavior, communication overhead, and computational complexity of the proposed HFLDD. 
Extensive experimental results based on multiple public datasets demonstrate that when data labels are severely imbalanced, the proposed HFLDD outperforms the baseline methods in terms of both test accuracy and communication cost.

\end{abstract}


\section{Introduction}
\IEEEPARstart{I}{n} recent years, the rapid advancement of IoT devices and smart terminals has led to an exponential increase in global data volume \cite{yuan2024secure}. Mobile edge computing (MEC) provides an efficient solution by offloading data storage, analysis, and processing to the network edge, significantly improving efficiency and response times \cite{li2018learning,zhao2022energy}. 
Meanwhile, advances in deep neural networks have advanced tasks such as image classification \cite{krizhevsky2012imagenet, han2022survey}, object detection \cite{redmon2018yolov3, ren2016faster,wang2024end}, and semantic segmentation \cite{long2015fully}, making machine learning (ML) workloads a primary focus in MEC. 
To address challenges such as privacy leakage and high communication costs, Federated Learning (FL) has become an emerging machine learning paradigm and has attracted widespread attention \cite{cai2025comprehensive,han2025rethinking,yin2021comprehensive, li2025accelerating}.


  In the widely used FL-based method FedAvg \cite{mcmahan2017communication}, the server updates the global model by aggregating the local model parameters with simply weighted averaging. 
  This method performs well when dealing with IID data.
  However, researchers have found that the data among clients is often non-IID in practice\cite{kairouz2021advances}. 
  In such cases, if the server simply aggregates the parameters of the local model with weighted averaging, the drift of the local model may cause the global model to deviate from the global optimum \cite{li2019convergence}, leading to slower convergence of the global model\cite{wang2021field}.

To mitigate the impact of non-IID data on FL-based model training,
researchers have explored various approaches. At the algorithm level, efforts focus on improving loss functions \cite{karimireddy2020scaffold,li2020federated,li2021model,guo2025FedVLS,lu2023FedLMD,lee2022FedNTD,acar2021FedDyn,zhang2022FedLC} and aggregation methods \cite{tan2022adafed, chai2021fedat,ye2023feddisco,li2023fedlaw,rehman2023Ldawa,shi2025fedawa}. 
However, they cannot fully alleviate the impact of non-IID data on model training and may have poor performance when dealing with high statistical heterogeneity. 
At the system level, client clustering techniques are introduced to create a multicenter framework \cite{ghosh2020efficient, he2022improving, wang2021edge}. These techniques aim to group clients with similar data distributions together to form clusters. 
However, this approach primarily focuses on the similarity of the client and does not fully exploit the potential advantages of the heterogeneity of the client.
At the data level, data sharing is considered an effective strategy \cite{zhao2018federated, gu2022fedaux, tuor2021overcoming}. It mitigates the negative impact of non-IID data by sharing a small amount of local data. 
Although this method can improve the performance of the global model, the server typically lacks comprehensive knowledge of the client data distributions, making it difficult to obtain a uniformly distributed dataset. 
Moreover, this approach conflicts with the privacy protection principles of FL.

\textcolor{black}{
Recent work on dataset distillation shows promising potential to overcome restrictions on data sharing while protecting privacy \cite{10273632, song2023federated, liu2022meta, pi2023dynafed,jia2024unlocking, xiong2023feddm, xu2024flip, wang2024aggregation, yan2025fedvck}. }Dataset distillation\cite{wang2018dataset,nguyen2021dataset,zhao2020dataset, zhao2023dataset, cazenavette2022dataset} aims to achieve good generalization performance by training models on a synthesized dataset that is much smaller than the real dataset. 
It has been indicated that the privacy of the transmission of distilled data should not be worse than that of the transmission of model parameters\cite{zhou2020distilled}. 
Additionally, distilled data not only provides visual privacy, but also enhances the model's resistance to membership inference attacks \cite{dong2022privacy}.

Motivated by the above observations, we propose a new learning framework named HFLDD, integrating dataset distillation with hybrid federated learning (HFL). 
The main idea of the design is to partition clients into a few clusters where data labels are heterogeneous within each cluster but balanced across different clusters. Furthermore, within each cluster, the cluster members transmit their distilled data to a cluster head. The adoption of client clustering and dataset distillation in FL can reduce the total communication cost of model training, while compensating for data heterogeneity between clients and improving the performance of the global model.

The contributions of our work are as follows.

\begin{enumerate}
    \item We propose a novel framework, HFLDD, to alleviate the negative impact of non-IID data on model training. Specifically, clients are partitioned into heterogeneous clusters and construct approximately IID data among different clusters by integrating dataset distillation.

    \item {We theoretically analyze the convergence behavior, communication overhead, and computational complexity of the proposed HFLDD. Specifically, we derive a convergence bound and provide a closed-form expression for the communication cost associated with HFLDD. Additionally, we explicitly characterize the computational complexity for the server, the cluster head, and the cluster member, respectively.}

    \item Experimental results in public datasets demonstrate that the proposed HFLDD outperforms baseline methods in terms of both test accuracy and communication cost, especially when the client data labels are highly unbalanced.

\end{enumerate}


The remainder of this paper is organized as follows.
Section \ref{sec:relatework} introduces the related work. Section \ref{sec:Preliminaries} provides the preliminaries and problem statement. Section \ref{sec:method} presents the proposed HFLDD framework in detail. Section \ref{sec:analyis} provides a theoretical analysis on the convergence, communication cost, and computational complexity of HFLDD.
Section \ref{sec:experiments} conducts extensive experimental evaluations to validate the effectiveness of HFLDD. 
Section \ref{sec:discuss} provides a discussion. 
Finally, Section \ref{sec:conclusion} presents the conclusion.

\section{RELATED WORK}\label{sec:relatework}

{Non-IID data poses a significant challenge to FL. Typical non-IID scenarios are generally classified into three categories \cite{kairouz2021noniidscenary}: feature distribution skew, quantity skew, and label distribution skew. Among these, label distribution skew, can be further divided into quantity-based and distribution-based subtypes. To address the non-IID issues, researchers have explored solutions from the algorithm, system, and data levels. Specifically, algorithm level methods concentrate on improving loss functions and aggregation methods; system-level methods focus on optimizing the framework's topological structure; and data-level methods seek to improve sharing, augmentation and distillation of local data. The basic frameworks of FL include traditional FL, decentralized FL (DFL), and HFL. Each framework employs different methods to address the non-IID problem. }

\textcolor{black}{\subsection{Traditional FL on Non-IID data}}
In traditional FL, all clients train models locally and then transmit these local models to a central server for global aggregation. Various methods have been proposed since the introduction of FedAvg \cite{mcmahan2017communication} to address the non-IID issue. Approaches such as FedProx \cite{li2020federated}, SCAFFOLD \cite{karimireddy2020scaffold}, and MOON \cite{li2021model} improve local training by modifying loss functions or introducing representation regularization. Another line of work focuses on dynamically adjusting aggregation weights.
FedDisco \cite{ye2023feddisco} determines weights by jointly considering dataset scale and discrepancies between local and global category distributions, while FedLaw \cite{li2023fedlaw} adaptively refines aggregation weights through gradient descent on a proxy dataset at the server, enabling a more balanced and data-aware global aggregation.

\textcolor{black}{
Apart from these model-centric approaches, data-centric solutions have also been explored, including data augmentation \cite{yoon2021fedmix}, data selection \cite{cao2023knowledge}, and data sharing \cite{zhao2018federated}. Recently, dataset distillation has emerged as a promising solution for data-centric federated learning.
FedMK \cite{liu2022meta} proposes dynamic weight assignment and meta-knowledge sharing to accelerate convergence.
DynaFed \cite{pi2023dynafed} generates distilled data by matching the global model trajectory after each aggregation and further updates the global
 model based on these distilled data.
 FedDM \cite{xiong2023feddm} improves communication efficiency and test accuracy by constructing distilled data on each client to locally match the loss landscape of the original data. 
 FLiP \cite{xu2024flip} applied the principle of least privilege to ensure that clients only share the necessary knowledge, further protecting sensitive information.
 FedAF \cite{wang2024aggregation} enables clients to collaboratively learn distilled data and soft labels to improve model performance.
 FedVCK \cite{yan2025fedvck} condenses valuable client knowledge into distilled data with latent distribution constraints, improving non-IID robustness and communication efficiency.
}

\textcolor{black}{\subsection{Decentralized FL on Non-IID data}}
DFL can implement distributed learning from data distributed across multiple clients without the need for a central server. DESA \cite{huang2024DESA} employs synthetic anchor data to achieve implicit feature alignment and cross-client knowledge transfer without relying on a central coordinator. 
DFPL \cite{zhang2025dfpl} enables client collaboration through prototype exchange and integrates blockchain to support a decentralized architecture.
Furthermore, some research focused primarily on how to utilize the topological structure to improve the learning process. DFedPGP \cite{liu2024DFedPGP} leverages directed communication topologies and gradient push optimization to efficiently learn shared feature layers and personalized classifier heads.
D-Cliques\cite{bellet2022d} address label shift problems by constructing heterogeneous topological structures based on the underlying data distribution of each client, which may induce privacy leakage. To enhance privacy preservation, Ref. \cite{abebe2023optimizing} adopts a proxy-based method to learn the knowledge of local dataset labels based on a global dataset and to construct a heterogeneous topological structure for DFL. The proposed method has shown that when the neighbors of each client have heterogeneous data, it can improve the convergence of model training in addressing label distribution skew problems. This insight inspires the main idea of our method. It should be noted that the paradigm of fully distributed topological structure construction and learning suffers from a high communication overhead and node inconsistency.

\newcommand{\cmark}{\checkmark}%
\newcommand{\xmark}{\ding{55}}%

\begin{table*}[htbp!]
\centering
\caption{Comparisons of different FL frameworks on Non-IID data}
\label{tab_comparison}
\begin{tabular}{llccccc}
\toprule
\multirow{2}{*}{\textbf{Framework}} & 
\multirow{2}{*}{\textbf{Reference}} & 
\multirow{2}{*}{\textbf{Non-IID scenario}} &
\multicolumn{3}{c}{\textbf{Method}} &
\multirow{2}{*}{\textbf{Limitation}} \\

\cmidrule(lr){4-6} 
 &  &  & \begin{tabular}{@{}c@{}}\textbf{Algorithm }\\ \textbf{level}\end{tabular} & \begin{tabular}{@{}c@{}}\textbf{System}\\ \textbf{level}\end{tabular} & \begin{tabular}{@{}c@{}}\textbf{Data}\\ \textbf{level}\end{tabular}  &  \\

\midrule
\multirow{9}{*}{Traditional FL}
 & FedAvg\cite{mcmahan2017communication} & Quantity-based label distribution skew & \cmark  &  \xmark & \xmark  
 &  \multirow{9}{*}{\parbox{3.2cm}{\centering Sensitive to data heterogeneity; \\Server coordination required.}} \\
 & FedProx\cite{li2020federated} & Quantity-based label distribution skew & \cmark  &  \xmark & \xmark  &   \\
 &  SCAFFOLD\cite{karimireddy2020scaffold} & Distribution-based label distribution skew & \cmark  &  \xmark  & \xmark  &   \\
 &  MOON\cite{li2021model} & Distribution-based label distribution skew & \cmark  & \xmark  & \xmark   &   \\
 & FedDisco\cite{ye2023feddisco} & Label distribution skew & \cmark  &  \xmark &  \xmark & \\
 & FedLaw\cite{li2023fedlaw} & Distribution-based label distribution skew & \cmark  & \xmark &  \xmark & \\
 & FedMK\cite{liu2022meta} & Distribution-based label distribution skew & \xmark  & \xmark  & \cmark  & \\
 & DynaFed\cite{pi2023dynafed} & Distribution-based label distribution skew & \xmark  & \xmark &  \cmark & \\
 & FedDM\cite{xiong2023feddm} & Distribution-based label distribution skew & \xmark  & \xmark &  \cmark & \\
\midrule
\multirow{5}{*}{Decentralized FL}
 & DESA\cite{huang2024DESA}  & Feature distribution skew & \cmark &  \xmark  &  \cmark
 &  \multirow{5}{*}{\parbox{3.2cm}{\centering High communication overhead; \\Node inconsistency.}} \\
 & DFPL\cite{zhang2025dfpl} & Quantity-based label distribution skew & \cmark & \xmark  & \xmark   \\
 & DFedPGP\cite{liu2024DFedPGP} & Label distribution skew & \xmark  &  \cmark & \xmark   \\
 & D-Cliques\cite{bellet2022d} & Quantity-based label distribution skew & \xmark  & \cmark  & \xmark   \\
 & Ref.\cite{abebe2023optimizing} & Label distribution skew & \xmark & \cmark  & \xmark   \\
\midrule
\multirow{5}{*}{Hybrid FL}
 & HFML\cite{li2025hfml} 
 & Quantity-based label distribution skew 
 & \xmark  & \cmark  & \xmark  
 &  \multirow{5}{*}{\parbox{3.2cm}{\centering High communication overhead.}} \\
 & Semi-FL\cite{chen2020semifl} & Quantity-based label distribution skew & \xmark  & \cmark  & \xmark   \\
 & FedCluster\cite{chen2020fedcluster} & Distribution-based label distribution skew & \xmark  & \cmark  & \xmark   \\
 & Ref.\cite{lin2021semi} & Quantity-based label distribution skew & \cmark & \cmark  &  \xmark   \\
 & FedSeq\cite{chen2023fedseq} & Distribution-based label distribution skew & \xmark  & \cmark  &  \xmark   \\
\bottomrule
\end{tabular}
\end{table*}

\textcolor{black}{\subsection{Hybrid FL on Non-IID data}}
Unlike traditional and decentralized FL, HFL involves both client-to-server communication and client-to-client communication. 
Many existing approaches focus on system-level innovations by designing novel architectures. For example,
HFML \cite{li2025hfml} introduces a client–edge–cloud architecture that applies deep mutual learning at the client–edge layer for knowledge sharing. In \cite{lin2021semi}, a multistage HFL algorithm is developed to handle non-IID data by continuously adjusting the rounds of client-to-client communication and global aggregation cycles.
Another representative system-level direction is client clustering. Semi-FL \cite{chen2020semifl} clusters clients and adopts in-cluster sequential training, enabling neighboring clients to share models and reduce communication overhead. FedCluster \cite{chen2020fedcluster} accelerates convergence by clustering clients and performing cyclic training.
FedSeq \cite{chen2023fedseq} is based on client clustering and sequential training. It partitions clients into multiple clusters, with each cluster head communicating with the server. Within each cluster, clients form a ring topology for sequential training. The core idea of FedSeq is that the model in the cluster head is based on the training data of all the clients within this cluster, thereby improving the test accuracy.

Table \ref{tab_comparison} summarizes different FL frameworks on non-IID data. In comparison with the aforementioned works, we design a new HFL framework to mitigate the challenges of non-IID data from both system level and data level. By constructing heterogeneous topology among clients and incorporating dataset distillation, the proposed method can effectively compensate the data heterogeneity among clients and achieve a well-performed global model with low communication overhead.

\section{Preliminaries and Problem Statement}\label{sec:Preliminaries}

\subsection{Dataset Distillation}\label{subsec:KIP}
The main idea of dataset distillation is to transfer a large dataset to a small dataset, while preserving the essential characteristics of the original dataset. 
    Suppose that the original data set and the distilled data are, respectively, denoted by $\mathcal{T}=\left\{\left({x}_{i}, y_{i}\right)\right\}_{i=1}^{|\mathcal{T}|}$ and $\mathcal{S}=\left\{\left(\hat{{x}}_{j}, \hat{y}_{j}\right)\right\}_{j=1}^{|\mathcal{S}|}$, where $\mathrm{x}_{i}\in\mathbb{R}^{d}$ and $y_{i}\in \mathbb{R}^{n_c}$ are, respectively, the input data and the corresponding label of the $i$-th sample in the original dataset; $\hat{{x}}_{j}\in\mathbb{R}^{d}$ and $\hat{y}_{j} \in \mathbb{R}^{n_c}$ are, respectively, the input data and the corresponding label of the $j$-th sample in the distilled data; $|\mathcal{T}|$ and $|\mathcal{S}|$ represent the sizes of the original dataset and the distilled data, respectively. Generally, the size of $\mathcal{T}$ is considerably larger than $\mathcal{S}$, i.e., $|\mathcal{S}| \ll|\mathcal{T}|$. 

To realize dataset distillation, one of the most widely used methods is Kernel Inducing Points (KIP) \cite{nguyen2020dataset, nguyen2021dataset}, which is a first-order meta-learning algorithm based on kernel ridge regression (KRR). Specifically, the KIP method is to find a ditilled dataset $\mathcal{S}$ that is approximate to the target dataset $\mathcal{T}$ by minimizing the following loss function 
\begin{equation}
\mathcal{L}(\mathcal{S}) =\frac{1}{2}\left\| Y_\mathcal{T}-K_{X_tX_s}\left( K_{X_sX_s}+\lambda I 
\right) ^{-1}Y_\mathcal{S}\right\|^2,
\end{equation}
where $\lambda>0$ is a regularization parameter; $Y_\mathcal{T} \in \mathbb{R}^{|\mathcal{T}| \times n_c}$ and $Y_\mathcal{S} \in \mathbb{R}^{|\mathcal{S}|  \times n_c}$ are the labels of $\mathcal{T}$ and $\mathcal{S}$, respectively; ${K}_{X_s X_s}\in \mathbb{R}^{|\mathcal{S}|\times |\mathcal{S}|}$ and $K_{X_tX_s}\in \mathbb{R}^{|\mathcal{T}|\times |\mathcal{S}|}$ are two kernel matrices defined as $(K_{X_s X_s})_{ij}=k\left(\hat{x}_{i}, \hat{x}_{j}\right)$ and $(K_{X_tX_s})_{ij}=k\left({x}_{i}, \hat{x}_{j}\right)$, where $k(\cdot)$ denotes the kernel function. KIP is highly efficient for the distillation of datasets because it involves only first-order optimization.

\subsection{Federated Learning}
Consider a classic FL scenario with $N$ participants collaboratively training a global model. Each participant possesses a local dataset, represented as 
$\mathcal{D}_i=\left\{ \left( X_i,\mathrm{y}_i \right) \right\}, i=1,\cdots,N$, where $X_i\in \mathbb{R}^{n_i \times d}$ and $\mathrm{y}_i\in \mathbb{R}^{n_i\times n_c}$ denote the tuple of input data points and their corresponding labels;
$n=\sum_{i=1}^N{n_i}$, where $n_i=|\mathcal{D}_i|$ is the number of samples in $\mathcal{D}_i$. The objective of collaborative learning is to solve the following distributed parallel optimization problem
\begin{equation}\label{eq:global_loss}
\underset{\omega}{\min}f\left( \omega \right) \triangleq 
\underset{\omega}{\min}\sum_{i=1}^N{\frac{n_i}{n}f_i\left( \omega \right)},
\end{equation}
where $\omega$ is the set of model parameters.  
The term $f_i(\omega)$ is defined as $\mathbb{E}_{\xi _i\sim 
\mathcal{D}_i}\mathcal{F}_i\left( \omega, \xi _i \right)$, where $\xi_i$ is a mini batch of data from $\mathcal{D}_i$, and $\mathcal{F}_i\left( \omega, \xi _i \right)$ is the loss function associated with $\xi_i$ and $\omega$.

In the process of training, each participating client trains the local model based on their local dataset and then transmits the local models to a server for global model aggregation. Specifically, in round $t$, client $i$ trains its local model $\omega_i^t$ by minimizing the local loss function $f_i\left( \omega \right)$ using the gradient descent method. Subsequently, the server updates the global model $\omega^t$ through aggregation rules, e.g., FedAvg. If client data are IID, the above method can provide the optimal solution of \eqref{eq:global_loss}. If the client data are non-IID, the obtained global model may be drifted from the optimum. The communication cost of the above process comes from the local model upload and the global model download.
By referring to \cite{zhou2020distilled}, the communication cost of the widely used FedAvg is

\textcolor{black}{
\begin{equation}
\mathcal{C}_{\text{FedAvg}}\,\, =\,\,N\cdot   \left( 2T-1 \right) \cdot |\omega|\cdot B_1,
\label{eq3}
\end{equation}}
where $T$ denotes the rounds of model communication, $|\omega|$ denotes the size of the model parameters, and $B_1$ denotes the number of bits consumed for transmitting each parameter.

\begin{table}[!b] 
\centering
\caption{Important Notations}
\label{tab:notation}
\begin{tabular}{cccc}
   \cmidrule[1.0pt](lr){1-4}
   Symbol & \multicolumn{3}{l}{Description} \\
   \cmidrule(lr){1-4}
    $N$ & \multicolumn{3}{l}{number of clients} \\
    $H$ & \multicolumn{3}{l}{number of heterogeneous clusters} \\
    $K$ & \multicolumn{3}{l}{number of homogeneous clusters after K-Means} \\
    $\mathcal{H}_{e}$ & \multicolumn{3}{l}{set of heterogeneous clusters} \\
    $\mathcal{H}_{e,h}$ & \multicolumn{3}{l}{set of clients in the $h$-th heterogeneous cluster} \\
    $\mathcal{H}_{o}$ & \multicolumn{3}{l}{set of homogeneous clusters} \\
    $\mathcal{H}_{o,k}$ & \multicolumn{3}{l}{set of clients in the $k$-th homogeneous cluster} \\
    $\mathcal{D}_i$ & \multicolumn{3}{l}{local dataset on client $i$}\\
    $\tilde{\mathcal{D}_i}$ & \multicolumn{3}{l}{distilled data on client $i$}\\
    $\mathcal{D}_g$ & \multicolumn{3}{l}{global dataset for pre-train}\\
    $N_c$ & \multicolumn{3}{l}{total number of label classes}\\
    $C$& \multicolumn{3}{l}{number of label classes in local dataset of each client} \\
    $a_h$& \multicolumn{3}{l}{cluster head in the $h$-th heterogeneous cluster} \\
    $\mathcal{A}$& \multicolumn{3}{l}{set of cluster head} \\
     $\mathcal{D}_h$ & \multicolumn{3}{l}{\textcolor{black}{hybrid dataset on the $h$-th cluster head}}\\
    $S_i$ & \multicolumn{3}{l}{\textcolor{black}{soft labels generated by client $i$}}\\
    $M$ & \multicolumn{3}{l}{similarity matrix} \\
    $\omega^t$ & \multicolumn{3}{l}{global model at communication round $t$}\\
    ${\omega}_{h}^{t}$ & \multicolumn{3}{l}{local model on the $h$-th cluster head at communication round $t$} \\
    $\xi _{h}^{t,k}$& \multicolumn{3}{l}{a mini batch of data sampled from $\mathcal{D}_h$} \\
    $\eta$ & \multicolumn{3}{l}{learning rate}\\
    $E_1$& \multicolumn{3}{l}{number of epochs in pre-train at label knowledge collection stage} \\
    $E_2$& \multicolumn{3}{l}{number of epochs in local training at model training stage} \\
     $B_1$& \multicolumn{3}{l}{number of bits consumed for transmitting one model parameter} \\
    $B_2$& \multicolumn{3}{l}{number of bits consumed for transmitting one data sample} \\
    $T$ & \multicolumn{3}{l}{number of model communication rounds} \\
   \bottomrule
\end{tabular}
\end{table}


\begin{figure*}[!htbp]
\centering
\includegraphics[width=0.7\textwidth]{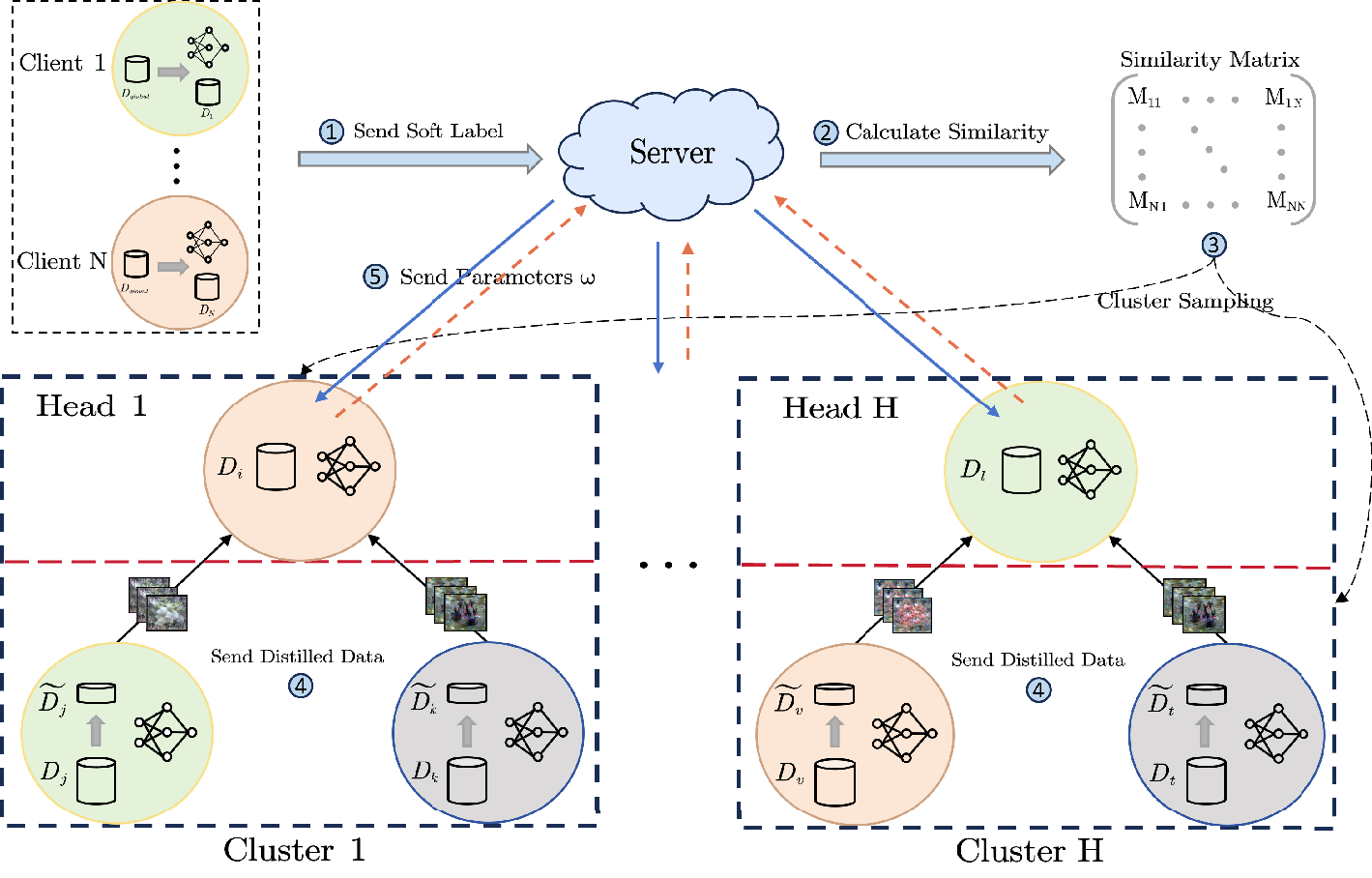}
\caption{\textcolor{black}{Overall pipeline of HFLDD. Different colors represent different label classes. Each client first performs local training and uploads soft labels to the server. The server conducts heterogeneous client clustering based on the similarity matrix. Within each heterogeneous cluster, all clients except the cluster head perform dataset distillation and transmit the distilled data to the corresponding cluster head. The server and cluster heads then collaboratively train the model.}} 
\label{fig:overall}
\end{figure*}

\subsection{Problem Statement}

In this paper, we consider a typical scenario of distributed learning with non-IID data, i.e., label distribution skew \cite{li2022federated}. Our focus lies on addressing the issue of quantity-based label imbalance, where each client owns data with a fixed number of label classes. 
Supposing $N_c$ denotes the total number of label classes, each client owns data with $C$ different labels. Data labels among different clients could overlap. If $C=N_c$, the data labels among all clients are the same, which can be regarded as quantity-based label balance. If $C<N_c$, the data labels among all clients are imbalanced. If $C=1$, it represents the worst-case scenario of quantity-based label imbalance, where the local dataset on each client contains only a single label class. The quantity-based label imbalance has been verified to greatly decrease the performance of model training \cite{li2020federated, geyer2017differentially}.

\textcolor{black}{The assumption of a fixed number of label classes per client is adopted primarily to provide a controlled and reproducible experimental setting, which is widely used in FL research for systematic evaluation of non-IID effects. In practical FL systems, however, clients may possess data from different numbers of label classes due to diverse user behaviors and data collection patterns. The proposed framework does not rely on clients having identical numbers of label classes. Instead, it leverages similarity in label knowledge inferred from soft labels to group clients and construct heterogeneous clusters, allowing it to naturally accommodate more general and realistic scenarios with arbitrary and uneven label distributions across clients.}


\section{The Proposed HFLDD}\label{sec:method}
In this section, we will first introduce the key idea of the proposed framework HFLDD. Then, we will introduce the details of each part.

\subsection{Overview}

 The key idea of HFLDD is to construct heterogeneous client clusters, where the data labels among different clusters are balanced and approximate the global distribution. The server can communicate with each cluster to obtain a global model similar to that achieved under IID data conditions. 

 \textcolor{black}{The overall pipeline of the proposed HFLDD framework is illustrated in Fig. \ref{fig:overall}, which consists of four key stages: 1) label knowledge collection, 2) heterogeneous client clustering, 3) IID dataset generation, and 4) model training. Each stage is associated with a specific set of operations, as described below:
\begin{enumerate}
    \item \textbf{Label Knowledge Collection (Step 1)}: Each client independently trains a local model using its private non-IID dataset. These locally trained models are then used to infer soft labels on a globally available auxiliary dataset. The resulting soft labels are uploaded to the server. This stage allows the server to gather label distribution information without accessing the raw private data, thus preserving client privacy.
\item \textbf{Heterogeneous Client Clustering (Steps 2–3)}: The server computes pairwise similarities between clients based on their soft labels and groups the clients with similar label distributions into homogeneous clusters. To enhance diversity, the server then constructs heterogeneous clusters by randomly sampling one client from each homogeneous cluster, and selects one client from each heterogeneous cluster as the cluster head. The clustering results are distributed to all clients.
\item  \textbf{IID Dataset Generation (Step 4)}: Within each heterogeneous cluster, all clients, except the cluster head, perform dataset distillation to generate distilled data that approximates their local data distribution. These distilled data are sent to the cluster head. The head aggregates all received distilled data, including its own data, to form an approximately IID hybrid dataset. This step helps mitigate statistical heterogeneity before training begins.
\item  \textbf{Model Training (Step 5)}: The server distributes the current global model to all cluster heads. Each head updates the model using the hybrid dataset and sends the updated model back to the server. The server aggregates the updated models using standard FL techniques such as FedAvg. This process is repeated for multiple rounds until model convergence.
\end{enumerate}}

\begin{figure*}[ht]
\centering
\includegraphics[width=0.7\textwidth]{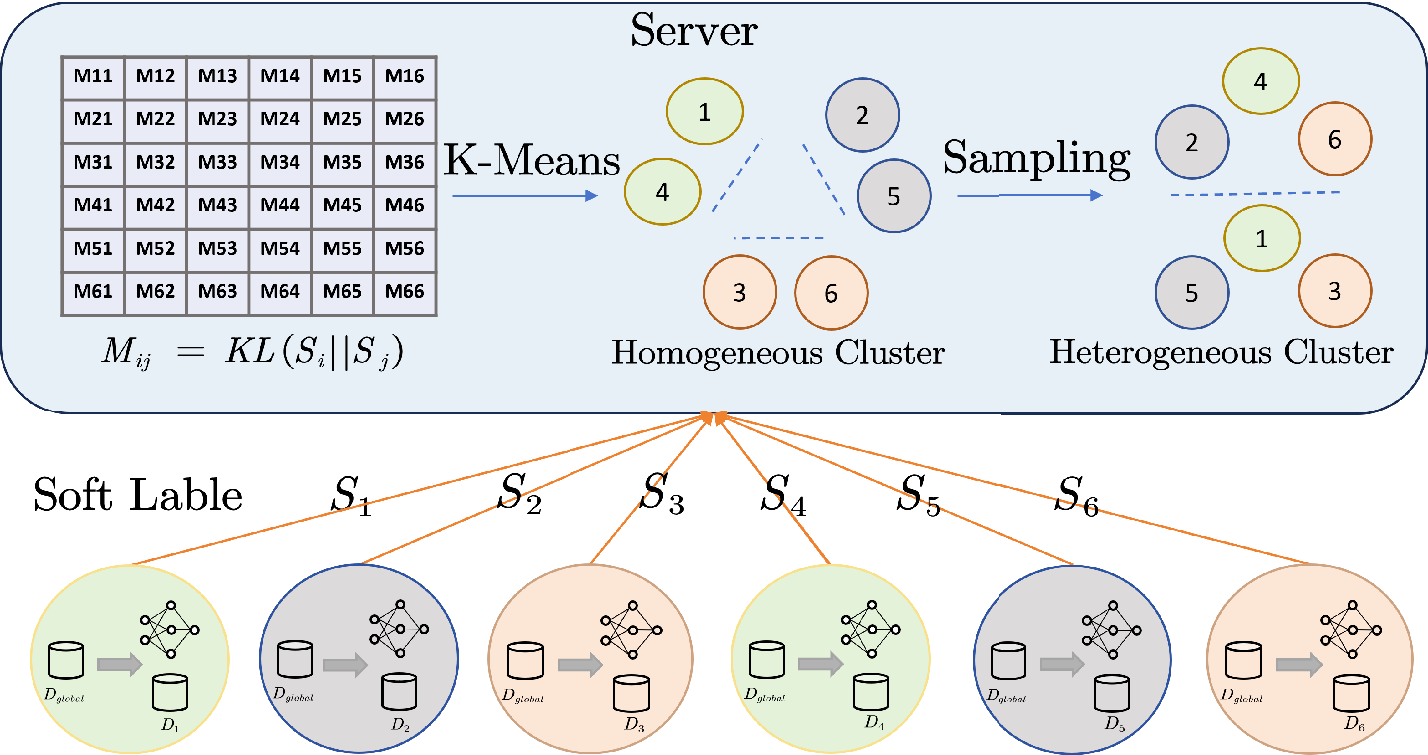}
\caption{\textcolor{black}{An overview of heterogeneous client clustering. Each circle represents a client, with different colors representing different data labels. Initially, the server collects the soft labels uploaded by clients and calculates the similarity matrix. The server groups the clients into homogeneous clusters through K-Means. Then, heterogeneous clusters are constructed by randomly sampling clients from different homogeneous clusters, ensuring label diversity within each heterogeneous cluster.}} 
\label{fig:cluster}
\end{figure*}

\subsection{Details}
The details of the implementation of each part in HFLDD are introduced in this subsection. Table \ref{tab:notation} summarizes the important notations.

\subsubsection{Label knowledge collection}
Each client first trains a local model on their own dataset $\mathcal{D}_i$. Since the trained local model is driven by the client dataset, it learns relevant knowledge about the local data. Subsequently, we introduce a finite and globally available dataset $\mathcal{D}_g=\{\left( X_g,y_g \right)\}$ \footnote{\textcolor{black}{The global dataset $\mathcal{D}_g$ used for soft labels generation can be a small, publicly available dataset related to the same task but it is not necessarily representative of the client data.}}, where $X_g\in \mathbb{R}^{|\mathcal{D}_g| \times d}$, $y_g\in \mathbb{R}^{|\mathcal{D}_g|\times N_c}$ and $|\mathcal{D}_g|$ is the number of samples in $\mathcal{D}_g$. The global dataset has a distribution different from the datasets of the clients and is used to generate soft labels for each client. 
Specifically, each client uses a globally available dataset as input for its locally pre-trained model to generate a two-dimensional tensor $S_i\in \mathbb{R}^{|\mathcal{D}_g|\times N_c}, \forall i=1,\cdots,N$.
This tensor represents the soft labels of the global dataset based on the local model of the client $i$. Let ${S}_{i,k}^{c}$ denote the probability that sample $k$ belongs to class $c$. It has been verified that the above soft labels can reflect the knowledge of the label of the local dataset \cite{abebe2023optimizing, abebe2024improving}. Therefore, after receiving the soft labels from the clients, the server obtains the label knowledge of each client without knowing the original dataset, which prevents data privacy leakage.

\begin{algorithm}[!hbpt]
	\renewcommand{\algorithmicrequire}{\textbf{Input:}}
	\renewcommand{\algorithmicensure}{\textbf{Output:}}
	\caption{ClusterSampling}
	\label{alg1}
	\begin{algorithmic}[1]
            \REQUIRE  $\mathcal{H}_{o}$, $K$
            \ENSURE Heterogeneous Cluster $\mathcal{H}_{e}$
        \STATE $h=1$
        \WHILE{ $\exists \mathcal{H}_{o,k} \neq \emptyset, \forall k=1,\cdots, K$ }
            \FOR{ $k=1,\cdots, K$}
                \IF{$\mathcal{H}_{o,k} \neq \emptyset$ }
                    \STATE select one client randomly from $\mathcal{H}_{o,k}$
                    \STATE add this client to $\mathcal{H}_{e,h}$
                    \STATE remove this client from $\mathcal{H}_{o,k}$
                \ENDIF
            \ENDFOR
              \STATE $\mathcal{H}_{e} \leftarrow \mathcal{H}_{e} \cup \{\mathcal{H}_{e,h}\}$
              \STATE $h=h+1$
        \ENDWHILE
        \RETURN $\mathcal{H}_{e}$
	\end{algorithmic}  
\end{algorithm}

\subsubsection{Heterogeneous Client Clustering}

Based on the soft labels of each client, the server calculates pairwise $KL$ divergences among these tensors to obtain the similarity matrix $M\in \mathbb{R}^{N\times N}$, where the $ij$-th element is calculated by
\begin{align}
M_{ij}&=KL\left( {S}_i||{S}_j \right) \nonumber\\
&=\frac{1}{|\mathcal{D}_g|}\left[ \sum_{k=1}^{|\mathcal{D}_g|}{\sum_{c=1}^{N_c}{\left( {S}_{i,k}^{c}\log \left( \frac{{S}_{i,k}^{c}}{{S}_{j,k}^{c}} \right) \right)}} \right].
\label{eq:similarity_matrix}
\end{align}
The element $M_{ij}$ indicates the difference in label knowledge between the local datasets of client $i$ and client $j$. 

Based on the similarity matrix, the server employs the K-Means algorithm to cluster all the
clients, by treating each row of the similarity matrix as a data point. 
This process groups clients with similar label knowledge into homogeneous clusters $\mathcal{H}_{o}=\left\{\mathcal{H}_{o,1}, \ldots, \mathcal{H}_{o,K}\right\}$. 
\textcolor{black}{While homogeneous clusters reveal similarity, our goal is to construct training groups with complementary label distributions. Heterogeneous clusters  $\mathcal{H}_{e}=\left\{\mathcal{H}_{e,1}, \ldots, \mathcal{H}_{e,H}\right\}$  are then formed by iteratively sampling across homogeneous groups to approximate inter-cluster label balance. } Fig. \ref{fig:cluster} illustrates the overall process of heterogeneous client clustering.  In this figure, 
each client's local data contains only one label, which makes the data among different clients are severe non-IID. After clustering, each homogeneous cluster contains only one label, while each heterogeneous cluster contains labels from all the clients.  The details of the sampling process are shown in Algorithm~\ref{alg1}.  \textcolor{black}{The value of $K$ controls the granularity of homogeneous grouping based on label-knowledge similarity. Its role is to ensure that clients with comparable statistical characteristics are first organized into representative similarity groups, which then serve as a structured pool for constructing heterogeneous clusters. Notably, some homogeneous clusters may contain more clients than others. After smaller clusters are exhausted, the algorithm continues to sample from the remaining larger clusters. In the case where only one homogeneous cluster remains, its remaining clients are sequentially assigned
to individual heterogeneous clusters. This will result in size imbalance, 
However, it can be mitigated by weighted aggregation and does not significantly impact convergence or model performance.}

Then, the server randomly selects one client within each heterogeneous cluster as the head for this cluster, and pushes the cluster information to each client. In this way, the clients construct a two-layer topology, as shown in Fig. \ref{fig:overall}

\subsubsection{IID dataset generation}
In each heterogeneous cluster, the cluster head acts as an agent for this cluster, conducts local model training, and communicates with the server. Each cluster member, except the head, sends a distilled data of much smaller size to the cluster head. In this work, clients employ the KIP method to obtain distilled data, as introduced in Section \ref{subsec:KIP}. For the $h$-th heterogeneous cluster, when the cluster head aggregates the distilled data from the entire cluster, it obtains a
hybrid dataset $\mathcal{D}_h$, which consists of its own local data and the distilled data received from its cluster members. We transform the problem into a learning process between the server and a number of cluster heads with approximately IID data. 

\subsubsection{Model Training}

As we mentioned above, after transforming our problem into distributed learning with IID data, we can apply the classical FL method, such as FedAvg\cite{mcmahan2017communication}, for model training. In the communication round $t$, the server broadcasts the global model parameters $\omega ^t$ to all heads of the connected cluster. Subsequently, the cluster head $h$ updates its local parameters to $\omega _{h}^{t}\,\,=\,\,\omega ^t$, and performs $E_2$ local updates in $\mathcal{D}_h$. At the $(\tau+1)$-th epoch of local update, the local model parameters of cluster head $h$ are updated as 
\begin{equation}
\omega _{h}^{t,\tau+1} = \omega _{h}^{t,\tau} - \eta\nabla F_h\left( \omega _{h}^{t,\tau},\xi _{h}^{t,\tau} \right),
\end{equation}
where $h=1,\cdots,H$, $\tau=0,\cdots, E_2-1$, $\xi _{h}^{t,\tau}$ denotes a randomly selected batch from $\mathcal{D}_h$, $\eta$ denotes the learning rate and $\nabla F_h\left(\cdot \right)$ denotes the gradient of the local loss function. Finally, the server aggregates the updated local models $\omega _{1}^{t,E_2},...,\omega _{H}^{t,E_2}$ from all the cluster heads to generate a new global model.

The whole process of HFLDD is illustrated in Algorithm \ref{alg2}.

\begin{algorithm}
	\renewcommand{\algorithmicrequire}{\textbf{Input:}}
	\renewcommand{\algorithmicensure}{\textbf{Output:}}
	\caption{HFLDD}
	\label{alg2}
	\begin{algorithmic}[1]
            \REQUIRE  $H$, $T$, $\mathcal{D}_i$, $\mathcal{D}_g$, $\eta$, $E_1$, $M$, $E_2$, $K$
            \ENSURE global model $\omega^T$
		\STATE Initialize server weights $\omega ^0$
  \STATE \textbf{Label knowledge collection:}
		  \FOR{$\text{client } i=1$ to $N$}
                \STATE send the global model $\omega ^0$ to client $i$
                \STATE local model $\omega _{i}^{0}\,\,\longleftarrow 
                \,\,\mathrm{local}\,\mathrm{training}\left( \mathcal{D}_i,\eta ,\omega ^0,E_1 \right)$
                \STATE predict soft labels $S_i$ by local model on dataset $\mathcal{D}_g$
                \STATE send $S_i$ to the server
            \ENDFOR
  \STATE \textbf{Heterogeneous client clustering:}  
            \FOR{$\text{client } i=1$ to $N$}
                \FOR{$\text{client } j=1$ to $N$}
                    \STATE $M_{ij} = \text{KL}(S_i || S_j)$
                \ENDFOR
            \ENDFOR
       \STATE $\mathcal{H}_{o}=\left\{\mathcal{H}_{o,1}, \ldots, \mathcal{H}_{o,K}\right\} \,\, \longleftarrow \,\, \text{K-Means}(K, M)$
        \STATE $\mathcal{H}_{e}\,\,\longleftarrow \,\,\mathbf{ClusterSampling}(K, \mathcal{H}_{o})$
        \STATE $\mathcal{A}=\emptyset$
             \FOR{ $h=1,\cdots, H$}
             \STATE select cluster head $a_h$ randomly from $\mathcal{H}_{e,h}$
             \STATE $\mathcal{A} \leftarrow \mathcal{A} \cup \{ a_h \}$
             \ENDFOR
  \STATE \textbf{IID dataset generation:}
		\FOR{$h=1,\cdots, H$}
                \FOR{client $i \in \mathcal{H}_{e,h}$ and $i \notin \mathcal{A}$}
                    \STATE $\tilde{\mathcal{D}_i}\gets KIP\left( \mathcal{D}_i\right) $
                    \STATE send distilled data $\tilde{\mathcal{D}_i}$ to $a_h$         
                \ENDFOR
                 \STATE $a_h$ combines its data with distilled data to form $\mathcal{D}_h$
            \ENDFOR
   \STATE  \textbf{Model training:}  
            \FOR{$t=1$ to $T$}
            \STATE broadcast the global model $\omega ^t$ 
                \FOR{$h=1,\cdots, H$ } 
                \STATE $\omega _{h}^{t}\,\,\longleftarrow 
                \,\,\mathrm{local}\,\mathrm{training}\left(\mathcal{D}_h,\eta ,\omega ^t,E_2 \right)$
                \ENDFOR
                \STATE $\omega ^{t+1}\gets aggregate(\omega _{1}^{t}, \ldots, \omega _{H}^{t}) $
            \ENDFOR
	\end{algorithmic}  
\end{algorithm}

\section{Performance Analysis}\label{sec:analyis}

\subsection{Convergence  Analysis}

\textcolor{black}{In HFLDD, the global training process is organized around a set of cluster heads. Each cluster head maintains a hybrid dataset and performs model training independently. The convergence behavior of HFLDD can be analyzed by extending the standard convergence theory of FedAvg with non-IID data. HFLDD maintains the following foundational assumptions commonly adopted in FL:}

\textcolor{black}{
\textbf{Assumption 1 ($L$-smoothness) :} For any model parameters $\upsilon $ and $\omega$, 
\begin{equation}
 ||\nabla f_{h}(\omega) - \nabla f_{h}(\upsilon)||^2\leq L ||\omega - \upsilon||^2, \forall h = 1, \dots, H.
\end{equation}
}
\textcolor{black}{
\textbf{Assumption 2 ($\mu$-strong convexity):} For any model parameters $\upsilon$ and $\omega$,
\begin{equation}
f_h(\upsilon) \geq f_h(\omega) + (\upsilon - \omega)^T \nabla f_h(\omega) + \frac{\mu}{2} ||\upsilon - \omega||^2, \forall h = 1, \dots, H.
\end{equation}
}
\textcolor{black}{
\textbf{Assumption 3 (Unbiased stochastic gradients):} 
Let $\xi_h \sim \mathcal{D}_h$ be a mini-batch sample from the hybrid dataset of cluster $h$, the stochastic gradient satisfies:
\begin{equation}
\mathbb{E}_{\xi_h \sim \mathcal{D}_h} \left[  \nabla f_{h}(w, \xi_h)\right]= \nabla f_{h}(w)
\end{equation}
\begin{equation}
\mathbb{E}_{\xi_h \sim \mathcal{D}_h} \left[ || \nabla f_{h}(w, \xi_h) - \nabla f_{h}(w)||^2 \right] \leq \sigma_h^2
\end{equation}
}
\textcolor{black}{
\textbf{Assumption 4 (Gradient Boundedness):} The expected squared norm of stochastic gradients is uniformly bounded: 
\begin{equation}
\mathbb{E}_{\xi_h \sim \mathcal{D}_h} \left[ || \nabla f_{h}(w, \xi_h) ||^2 \right] \leq G^2, ~\forall h = 1, \dots, H.
\end{equation}
To quantify the degree of data heterogeneous across clusters and the approximation error of dataset distillation, we further make the following assumptions:
}

\textcolor{black}{
\textbf{Assumption 5 (Bounded heterogeneity across clusters):} Let $\bar{\mathcal{D}}_h$ denote the full dataset containing the raw data of all the clients in cluster $h$. Let the optimal solution of the global model be $\omega^* \triangleq \arg\min_{\omega} f(\omega)$, the minimum global loss be $f^* \triangleq f(\omega^*)$, and the minimum local loss at cluster $h$ based on $\bar{\mathcal{D}}_h$ be $f_{h|\bar{\mathcal{D}}_h}^*$, then
\begin{equation} 
|f^*-\sum_{h=1}^H \mathrm{w}_h f_{h|\bar{\mathcal{D}}_h}^* |\leq \Gamma_{cluster}, 
\end{equation}
where $\mathrm{w}_h$ denotes the aggregation weight of local model at cluster $h$. 
}

\textcolor{black}{
\textbf{Assumption 6 (Bounded distillation  error):}  The approximation error between the empirical loss on the dataset with distillation and that without distillation at cluster $h$ is bounded: 
\begin{equation}
| f_{h|\bar{\mathcal{D}}_h}^*-f_{h|{\mathcal{D}}_h}^* |\leq \Gamma_{distill}.
\end{equation}
}

\textcolor{black}{
Based on the above six assumptions, we have the following theorem with regards to the convergence of HFLDD. 
}

\textcolor{black}{
\begin{theorem}
Under Assumptions 1–6, if the learning rate follows a diminishing schedule $\eta_t = \frac{2}{\mu(\tau + t)}$, with $\tau = \max\left\{8  \frac{L}{\mu},\ E_2\right\} - 1$, and all cluster heads participate in each communication round, then after $T$ rounds of communication, the global model $\omega^T$ satisfies:
\begin{equation}\label{eq:convergece bound}
\begin{aligned}
\mathbb{E}[f(\omega^T)] - f^*  
&\leq \frac{L}{T + \tau} \left( \frac{2 Q}{\mu^2} + \frac{\tau + 1}{2} || \omega^0- \omega^* ||^2 
\right),
\end{aligned}
\end{equation}
where 
\begin{equation}
Q= \sum_{h=1}^H \mathrm{w}_h^2 \sigma_h^2 
+ 6L \left(\Gamma_{cluster}+\Gamma_{distill}\right)
+ 8(E_2 - 1)^2 G^2.
\end{equation}
\end{theorem}
}

\begin{proof}
\textcolor{black}{
In HFLDD, only the cluster heads participant in global model training. It can be regarded as standard FedAvg on a cluster-level hybrid dataset $\mathcal{D}_h$. Based on the convergence analysis of FedAvg under non-IID conditions, as shown in \cite{li2019convergence}, the convergence bound of HFLDD satisfies: 
}
\textcolor{black}{
\begin{equation}\label{eq:convergece bound_FedAVg}
\begin{aligned}
\mathbb{E}[f(\omega^T)] - f^*  
&\leq \frac{L}{T + \tau} \left( \frac{2 B}{\mu^2} + \frac{\tau + 1}{2} || \omega^0- \omega^* ||^2 
\right),
\end{aligned}
\end{equation}
where 
\begin{equation}\label{eq:B}
B= \sum_{h=1}^H \mathrm{w}_h^2 \sigma_h^2 + 6L\Gamma+ 8(E_2 - 1)^2 G^2
\end{equation}
}

\textcolor{black}{
The parameter $\Gamma=f^*-\sum_{h=1}^H \mathrm{w}_h f_{h|{\mathcal{D}}_h}^*$ quantifies the degree of non-IID across datasets ${\mathcal{D}}_h, h=1,\cdots, H$. Under the assumptions of bounded heterogeneity and distillation approximation error, we have
\begin{align} \label{eq:Gamma}
\Gamma &=\underbrace{f^* - \sum_{h=1}^H \mathrm{w}_h f_{h|{\bar{\mathcal{D}}_h}}^*}_{\text{cluster heterogeneity}}
+ \underbrace{\sum_{h=1}^H\mathrm{w}_h \left( f_{h|{\bar{\mathcal{D}}_h}}^*- f_{h|{{\mathcal{D}}_h}}^*\right)}_{\text{distillation error}}
\nonumber\\
&\leq \left|f^* - \sum_{h=1}^H \mathrm{w}_h f_{h|{\bar{\mathcal{D}}_h}}^* \right| + \sum_{h=1}^H  \mathrm{w}_h \left|  f_{h|{\bar{\mathcal{D}}_h}}^*- f_{h|{{\mathcal{D}}_h}}^*\right|\nonumber\\
&\leq \Gamma_{cluster}+\Gamma_{distill}
\end{align}
}
\textcolor{black}{
Combining Eq.\eqref{eq:B} and Eq.\eqref{eq:Gamma}, we can easily derive that $B\leq Q$. Straightforwardly, Eq. \eqref{eq:convergece bound} holds. This concludes the proof. 
}
\end{proof}

\textcolor{black}{Theorem 1 demonstrates that under Assumption 1-6, when using a diminishing leaning rate $\eta_t=\mathcal{O}(1/t)$, the expected global loss converges to the optimal value at a rate of $\mathcal{O}(1/t)$. Moreover, the convergence bound is closely related to the heterogeneity of the cluster and the distillation error. The higher the heterogeneity of the cluster or the distillation error, the larger the convergence bound.}

\subsection{Communication Cost}
    Now, we analyze the total communication cost of HFLDD. Considering that the initialization of the server model can be sent as a random seed \cite{zhou2020distilled}, we can ignore the communication cost of sending random seeds. \textcolor{black}{Thus, we only consider the resources consumed for uploading soft labels to the server at the stage of label knowledge collection, transmitting the distilled data within each cluster at the stage of IID dataset generation, and communication between the server and cluster heads at the stage of model training.}   

Let $a_h$ denote the head of the heterogeneous cluster $h$, and $\left| \tilde{\mathcal{D}_i} \right|$ denote the size of the distilled data on the client $i$. 
Suppose that transmitting each parameter incurs a communication overhead of $B_1$ (usually
32 bits), and transmitting each distilled data incurs a communication overhead of $B_2$ (1 × 8 bits or 3 × 8 bits for each grayscale or RGB pixel). \textcolor{black}{The bit length $B_2$ reflects the transmission cost of each image sample, and no additional compression beyond the inherent reduction via dataset distillation is applied.} For $N$ clients, the communication cost for uploading the soft labels is $N \cdot |\mathcal{D}_g| \cdot N_c \cdot B_1$. The total communication cost to transmit the distilled data within each cluster is $\sum_{h=1}^H{\sum_{\substack{i\in\mathcal{H}_{e,h}\\i\neq a_h}} {\left| \tilde{\mathcal{D}_i} \right|}}\cdot B_2$. In the model training stage, the total cost of model communication between the server and the heads of the cluster in $T$ rounds is $H \cdot |\omega| \cdot (2T-1) \cdot B_1$. Consequently, the total communication cost of HFLDD is 

\textcolor{black}{
\begin{align}\label{eq:communicationhfldd}
\mathcal{C}_{\text{HFLDD}}& =  N \cdot |\mathcal{D}_g| \cdot N_c \cdot B_1 + \sum_{h=1}^H{\sum_{\substack{i\in\mathcal{H}_{e,h}\\i\neq a_h}} {\left| \tilde{\mathcal{D}_i} \right|}}\cdot B_2\nonumber \\
& ~~~+ H \cdot |\omega| \cdot (2T-1) \cdot B_1 .
\end{align}
}

\subsection{Computational Complexity}
The computational complexities of HFLDD at the server, the cluster members and the cluster heads are separately analyzed.
\subsubsection{Server}
In HFLDD, the server is responsible for client clustering and global model aggregation. 
In the process of client clustering, the server needs to compute the similarity matrix $M$, which involves the computation of KL Divergence between any pair of clients' soft labels. As shown in Eq.~\eqref{eq:similarity_matrix}, for each pair of clients $i$ and $j$, the complexity of computing $M_{ij}$ over the global dataset $D_g$ is $\mathcal{O}(|D_g| \cdot N_c)$. In total, the complexity of constructing the similarity matrix $M$ is $\mathcal{O}(N^2 \cdot |D_g| \cdot N_c)$. Based on the similarity matrix $M$, the server conducts homogeneous client clustering using K-Means algorithm, which incurs a complexity of $\mathcal{O}(N_{iter} \cdot K \cdot N^2)$, where $N_{iter}$ is the number of iterations in K-means. 
In the process of global model aggregation, the server aggregates models from all $H$ cluster heads over $T$ communication rounds. Each aggregation involves weighted averaging $H$ local models with size being $|\omega|$, resulting in a total computational cost of $\mathcal{O}(T \cdot H \cdot |\omega|)$. 

\textcolor{black}{
\subsubsection{Cluster Member} Each cluster member performs pre-training and soft labels generation at the stage of label knowledge collection and dataset distillation at the stage of IID dataset generation. 
}
\textcolor{black}{
In the process of local pre-training, each client performs $E_1$ steps of model updates on its local dataset using SGD. Let $b_1$ denote the mini-batch size used in each step. The computational complexity of local pre-training per client is $\mathcal{O}(E_1 \cdot b_1 \cdot |\omega|)$.   After pre-training, each client generates soft labels by performing inference over the public dataset $\mathcal{D}_g$, which incurs an additional computational cost of $\mathcal{O}(|\mathcal{D}_g| \cdot |\omega|)$. Therefore, the computation cost per client at the label knowledge collection stage is $\mathcal{O}(E_1 \cdot b_1 \cdot |\omega| + |\mathcal{D}_g| \cdot |\omega|)$.
}
\textcolor{black}{
In the process of dataset distillation, each cluster member synthesizes distilled data via the KIP method, which involves repeated optimization over $R$ iterations. In each iteration, for cluster member $i$, $i \in \mathcal{H}_{e,h}$ and $i \notin \mathcal{A}$, the most computationally intensive step is the inversion of the regularized kernel matrix in kernel ridge regression, which incurs a cost of $\mathcal{O}(|\tilde{\mathcal{D}_i}|^3)$. Therefore, the complexity of dataset distillation at cluster member $i$ is $\mathcal{O}(R \cdot |\tilde{\mathcal{D}_i}|^3).$
}
\textcolor{black}{
\subsubsection{Cluster Head} At the stage of label knowledge collection, each cluster head follows the same pre-training and soft labels generation steps as cluster members, which incurs a computational complexity of $\mathcal{O}(E_1 \cdot b_1 \cdot |\omega| + |\mathcal{D}_g| \cdot |\omega|)$. At the stage of global model training, each cluster head additionally conducts local training on its hybrid dataset.
In each communication round, the cluster head $a_h$ performs $E_2$ steps of SGD on its hybrid dataset $\mathcal{D}_h$. Let $b_2$ denote the batch size. The computational complexity of local training per round on a single cluster head is $\mathcal{O}(E_2 \cdot b_2 \cdot |\omega|)$.
 For all $T$ rounds, the total complexity at the global model training stage is $\mathcal{O}(T \cdot E_2 \cdot b_2 \cdot |\omega|)$.
}

\section{Experimental Evaluation}\label{sec:experiments}
In this section, we first analyze the performance of the proposed HFLDD and compare it with baseline methods to demonstrate how our method can mitigate the impact of label distribution skewness. Subsequently, we compare the communication overhead of different methods. Finally, we investigate the influence of several key hyperparameters and model architectures on the performance of HFLDD.

\subsection{Experimental Setup}
\subsubsection{Datasets}
Our experiments are carried out on three widely used datasets: MNIST \cite{lecun1998gradient}, CIFAR10 \cite{krizhevsky2009learning}, and CINIC10 \cite{cinic10}. MNIST consists of 50,000 binary images of handwritten digits, which can be divided into 10 classes. CIFAR10 consists of 50,000 daily object images, which can also be divided into 10 classes. 
\textcolor{black}{
CINIC10, as an extended version of CIFAR10, comprises 270,000 images and retains the original 10 classes of CIFAR10. By incorporating downsampled images from ImageNet, CINIC10 significantly increases the complexity and diversity of the dataset.}
In our non-IID setting, we investigate scenarios with each client has 1, 2, or 10 of the ten classes. In our experiments, each dataset is divided into 80\% for training and 20\% for testing.

In addition to training and testing the local client dataset, we also need additional global datasets to generate soft labels for label knowledge collection. We randomly select 1,000 samples from other similar datasets and construct the corresponding global dataset.
For the MNIST experiment, the global dataset is constructed from the FMNIST dataset \cite{xiao2017fashion}. For the CIFAR10 and CINIC10 experiments, the global dataset is constructed from the Tinyimagenet dataset \cite{le2015tiny}.


\subsubsection{Models}
In our experiments, the applied models are all based on Convolutional Neural Network (CNN) \cite{krizhevsky2012imagenet}.
For the MNIST dataset, we use a shallow CNN consisting of two convolutional layers with ReLU activations and max pooling, followed by flattening and three fully connected layers. The model contains 44,426 parameters. For the CIFAR10 dataset, a deeper CNN is adopted with four convolutional blocks. Each block includes two convolutional layers with ReLU activations, batch normalization, and a max pooling layer. After flattening, a single fully connected layer is applied. The total number of parameters is 1,020,160.
\textcolor{black}{
For the CINIC10 dataset, the model consists of three repeated convolutional blocks, each comprising a convolutional layer with 128 filters, followed by an instance normalization layer, a ReLU activation function, and an average pooling layer. The total number of model parameters is 320,010.
}
The input for all models is $32 \times 32 \times 3$ image and the output is a 10-dimensional one-hot class label. It should be noted that the proposed HFLDD is not limited by the model architectures, and it can be applied to any model. In the experiments, we also test the performance of HFLDD using different models. 

\begin{figure*}[!b]
\centering
\subfloat[MNIST ($C=1$)]{\includegraphics[width=0.3\linewidth]{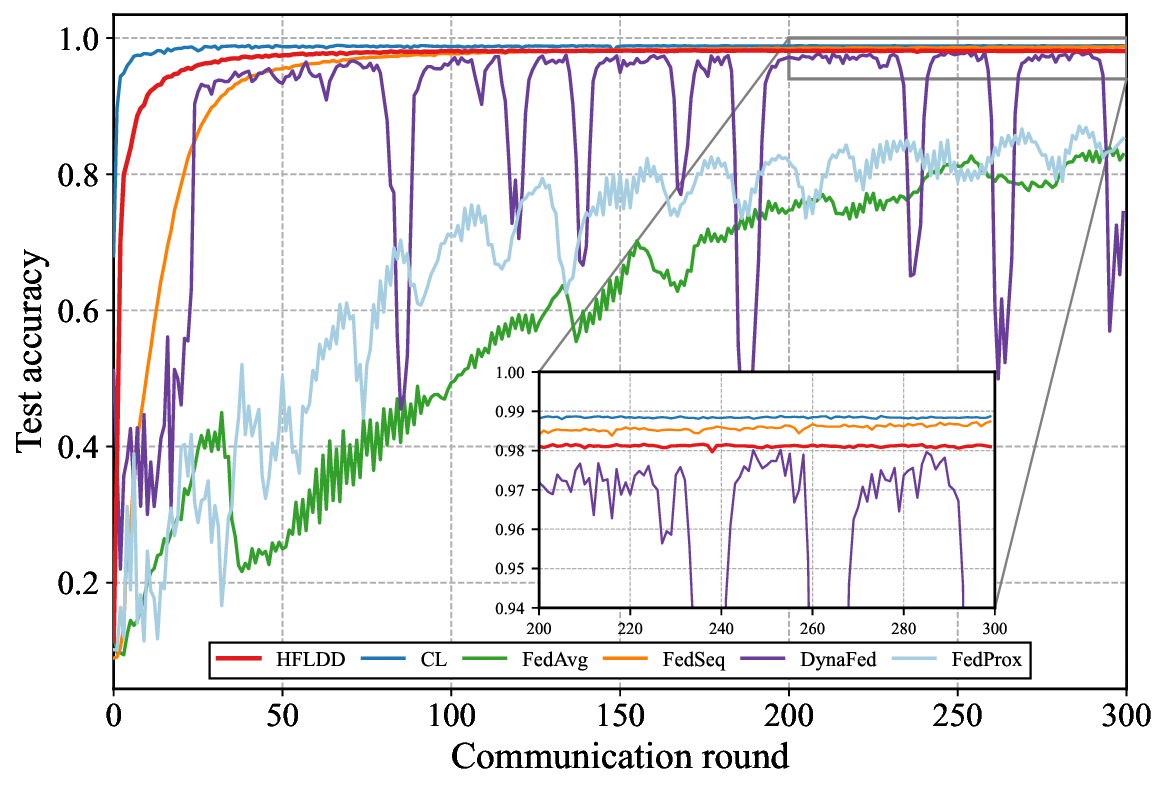}%
\label{subfig:accuracy1}}
\hfil
\subfloat[MNIST ($C=2$)]{\includegraphics[width=0.3\linewidth]{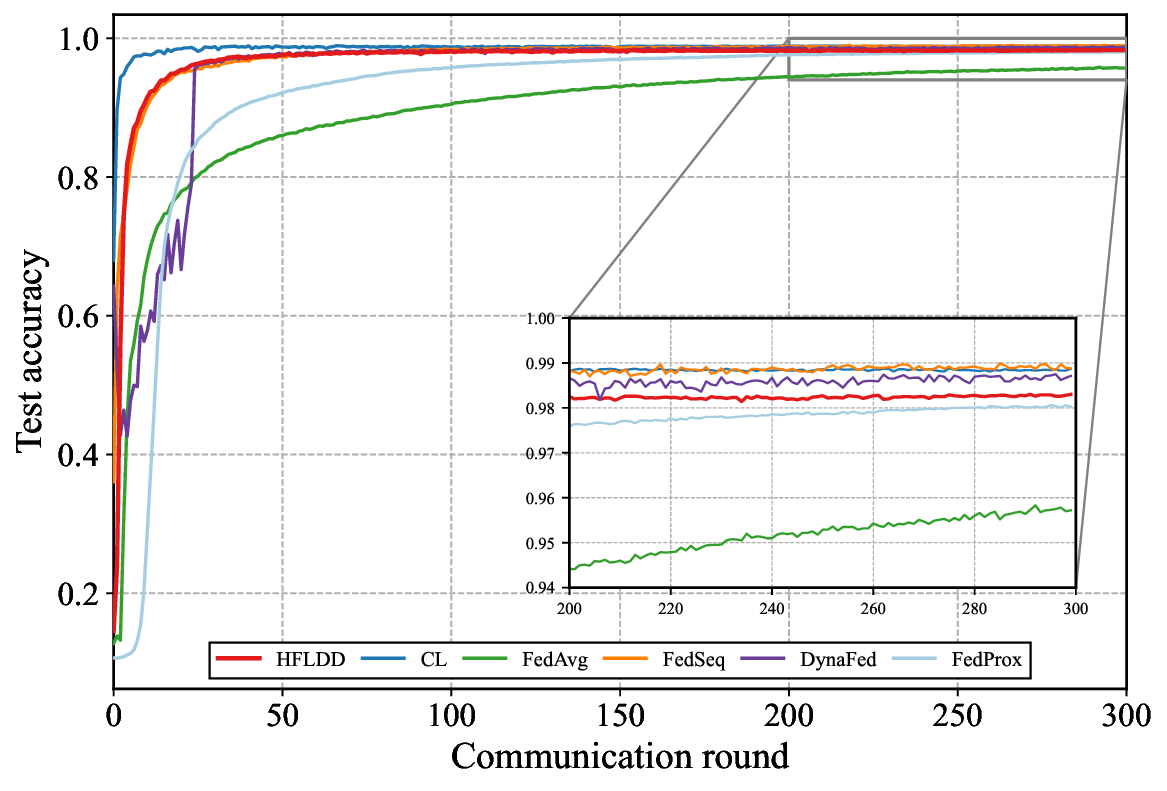}%
\label{subfig:accuracy2}}
\hfil
\subfloat[MNIST ($C=10$)]{\includegraphics[width=0.3\linewidth]{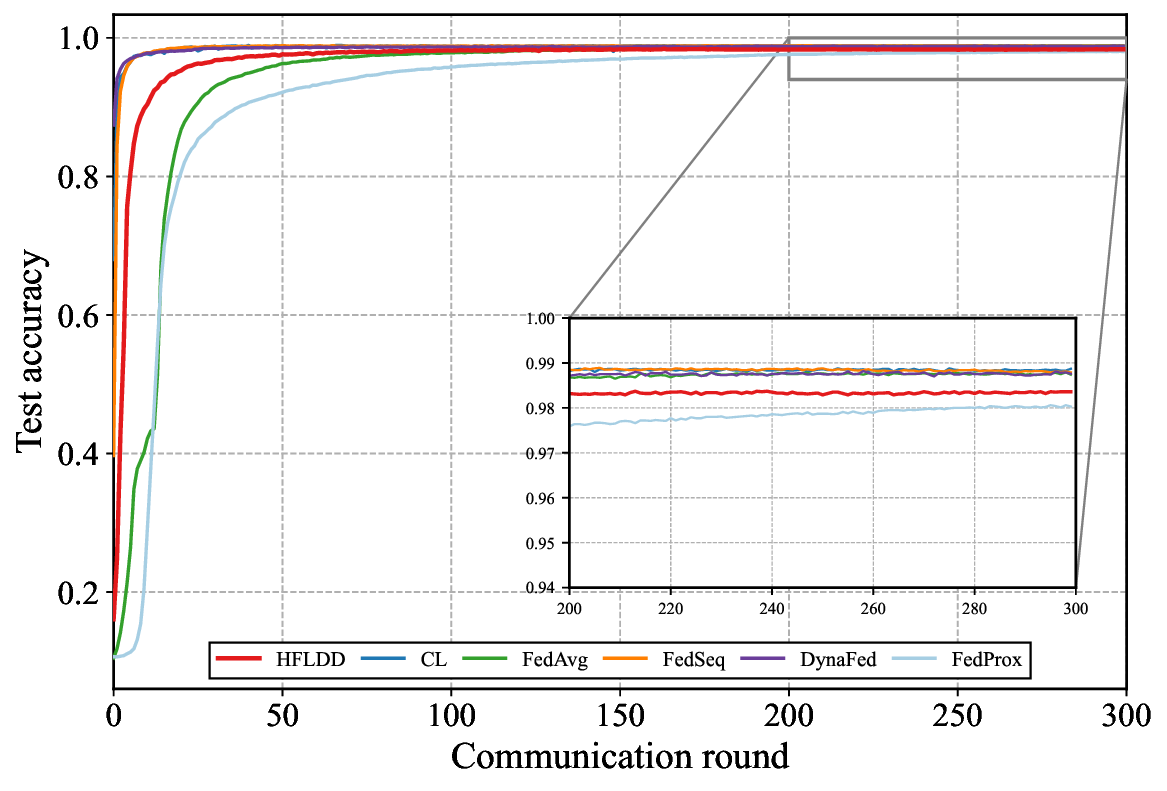}%
\label{subfig:accuracy3}}

\subfloat[CIFAR10 ($C=1$)]{\includegraphics[width=0.3\linewidth]{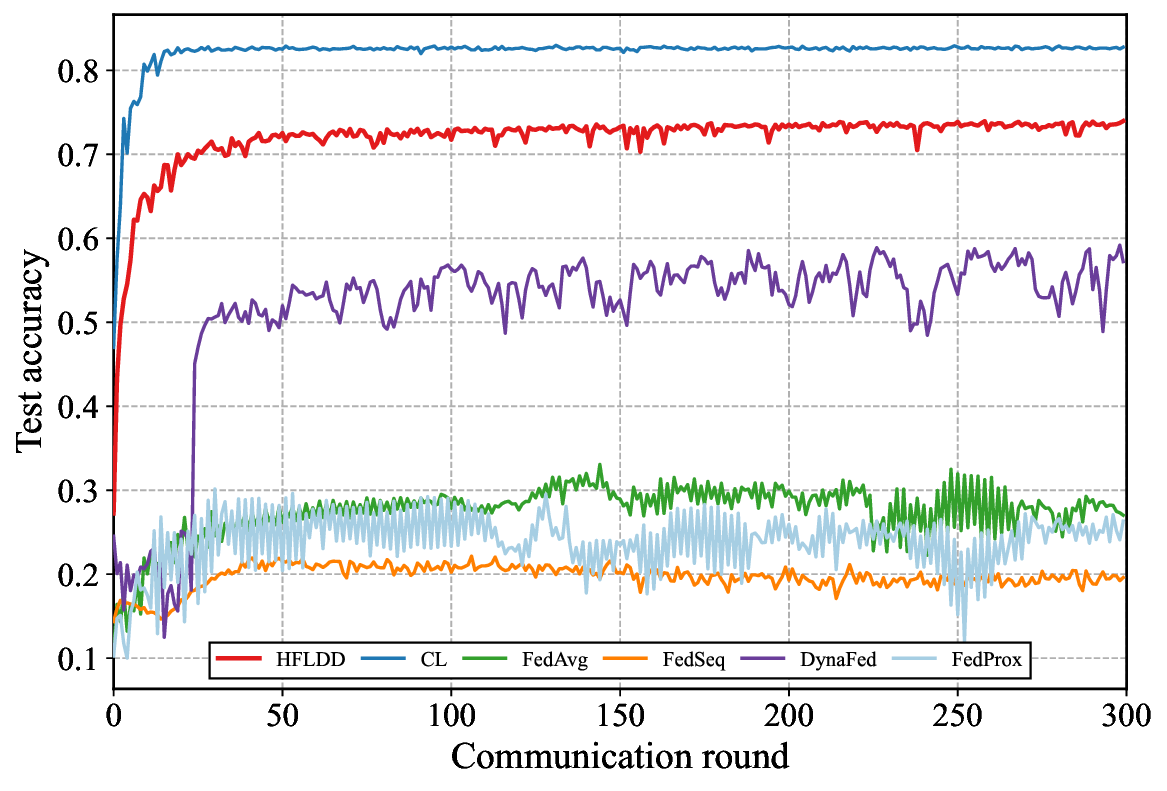}%
\label{subfig:accuracy4}}
\hfil
\subfloat[CIFAR10 ($C=2$)]{\includegraphics[width=0.3\linewidth]{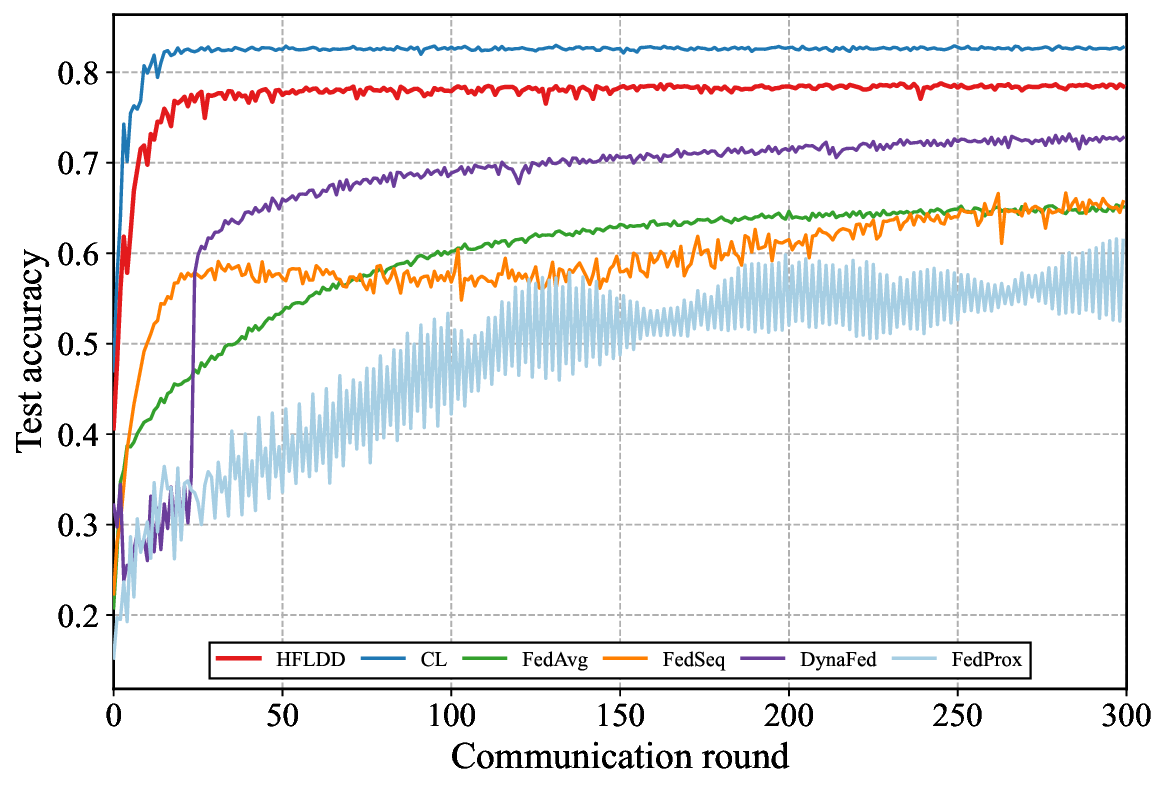}%
\label{subfig:accuracy5}}
\hfil
\subfloat[CIFAR10 ($C=10$)]{\includegraphics[width=0.3\linewidth]{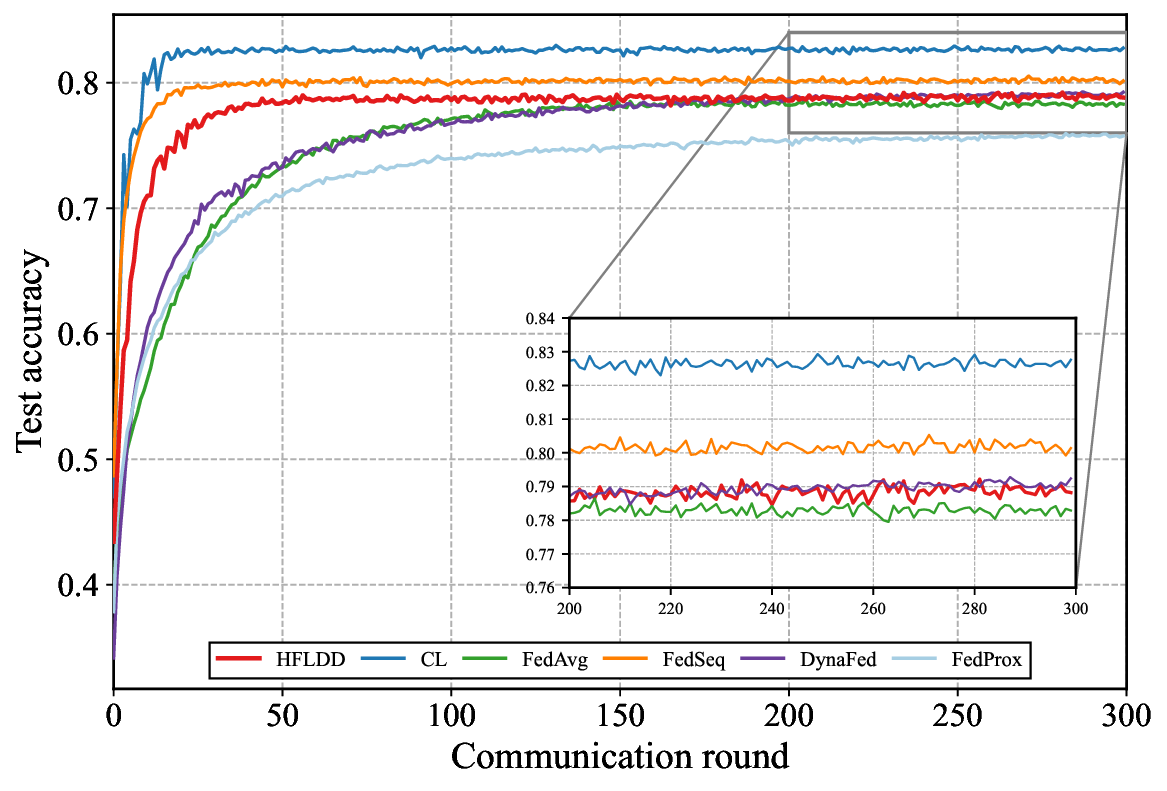}%
\label{subfig:accuracy6}}

\subfloat[CINIC10 ($C=1$)]{\includegraphics[width=0.3\linewidth]{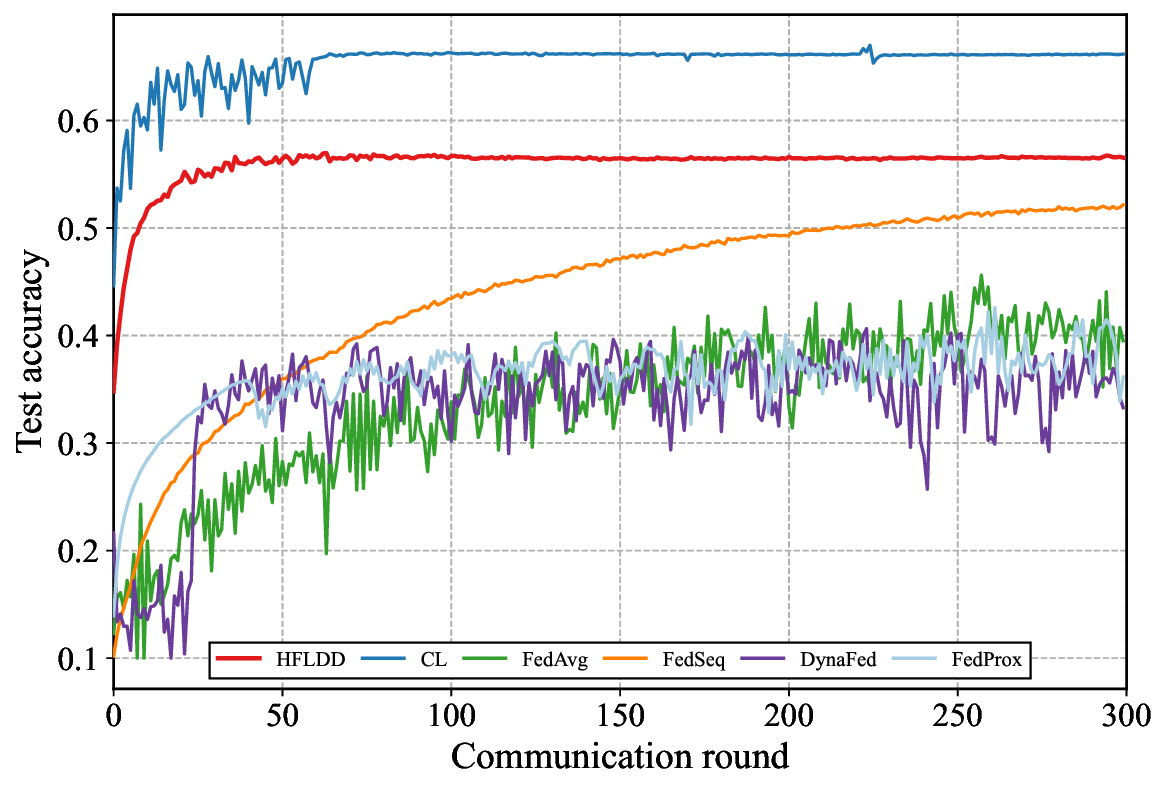}%
\label{subfig:accuracy7}}
\hfil
\subfloat[CINIC10 ($C=2$)]{\includegraphics[width=0.3\linewidth]{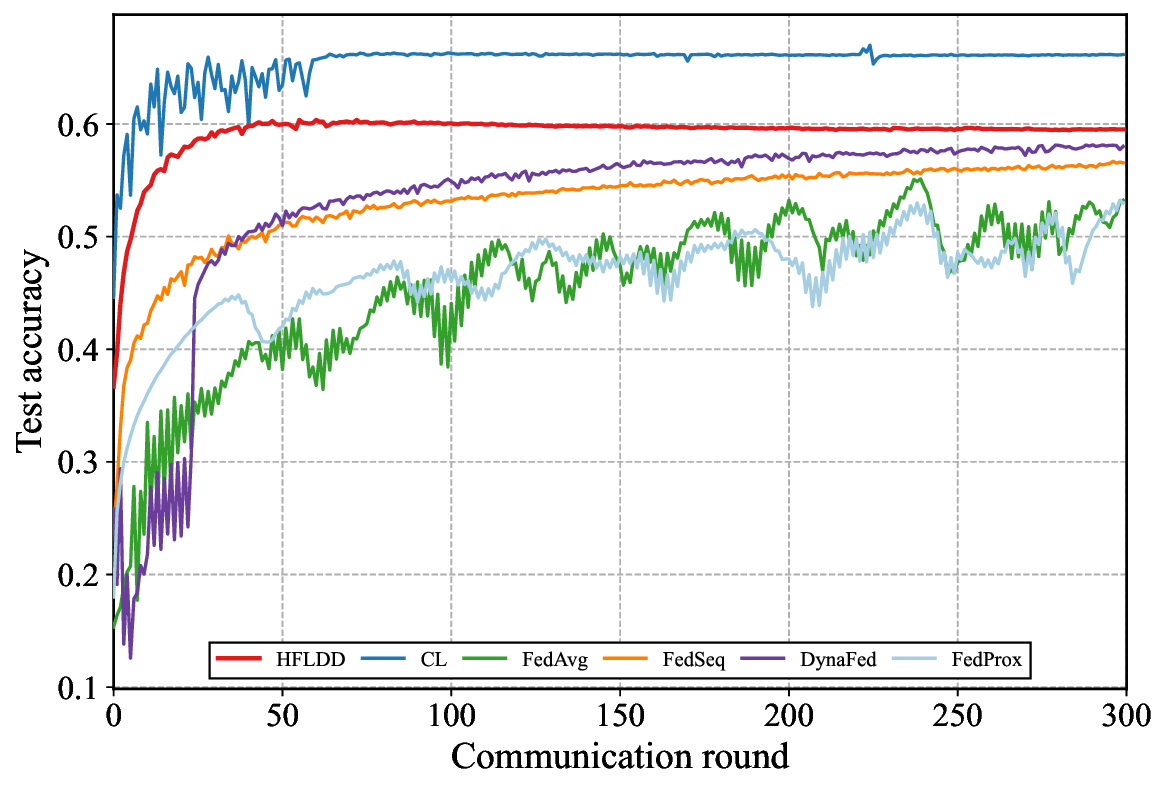}%
\label{subfig:accuracy8}}
\hfil
\subfloat[CINIC10 ($C=10$)]{\includegraphics[width=0.3\linewidth]{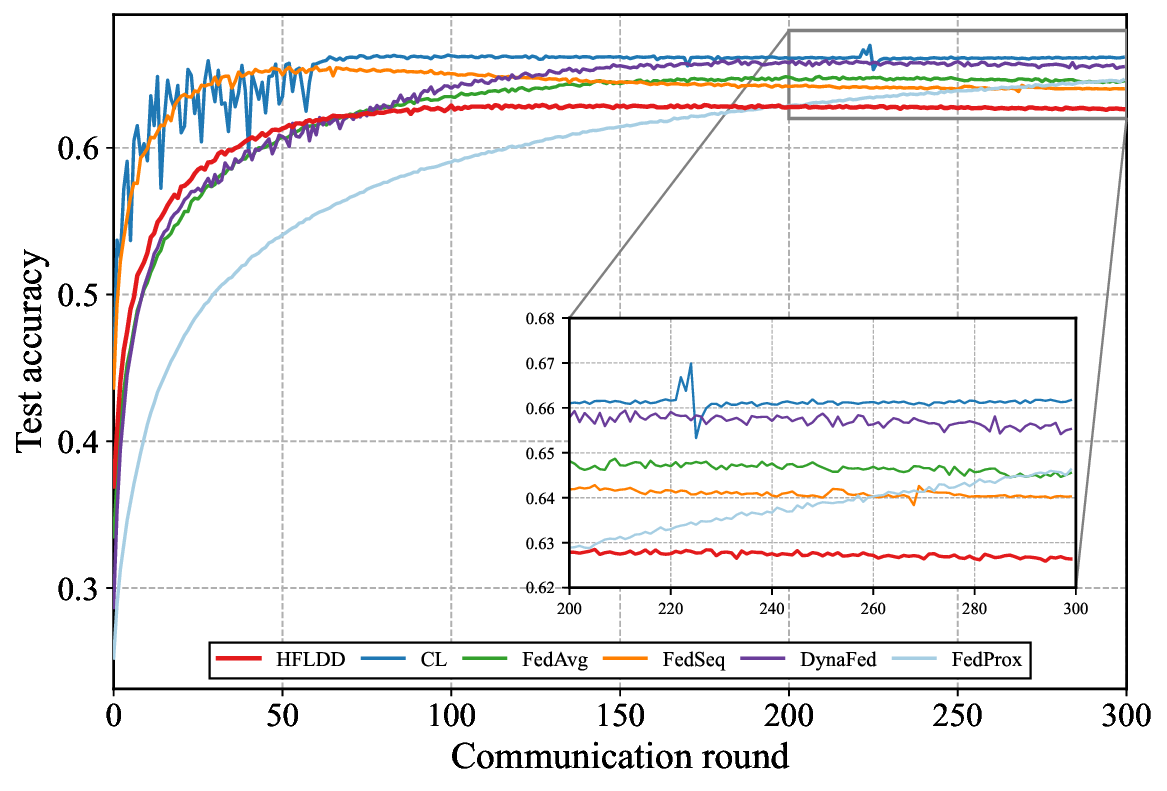}%
\label{subfig:accuracy9}}

\caption{
\textcolor{black}{
Learning curves of the global model on MNIST, CIFAR10 and CINIC10. 
}}
\label{fig:accuracy}
\end{figure*}

\begin{figure*}[b]
\centering
\subfloat[MNIST ($C=1$)]{\includegraphics[width=1.8in]{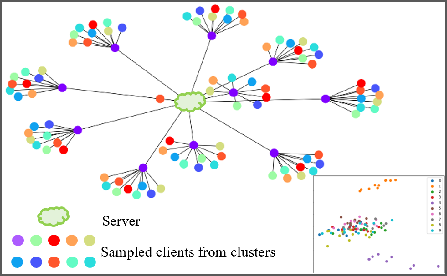}%
\label{subfig:topology111}}
\hfil
\subfloat[MNIST ($C=2$)]{\includegraphics[width=1.8in]{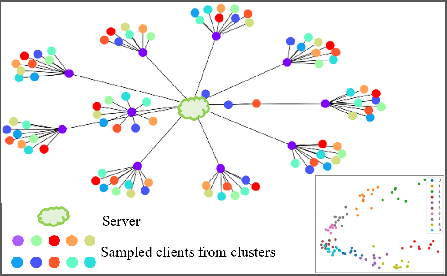}%
\label{subfig:topology222}}
\hfil
\subfloat[MNIST ($C=10$)]{\includegraphics[width=1.8in]{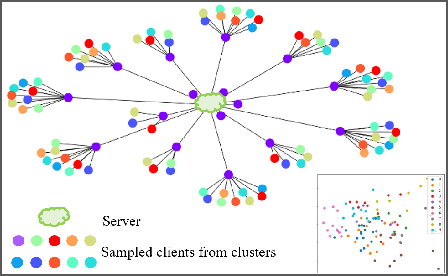}%
\label{subfig:topology333}}

\subfloat[CIFAR10 ($C=1$)]{\includegraphics[width=1.8in]{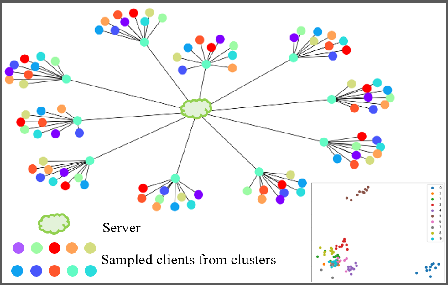}%
\label{subfig:topology444}}
\hfil
\subfloat[CIFAR10 ($C=2$)]{\includegraphics[width=1.8in]{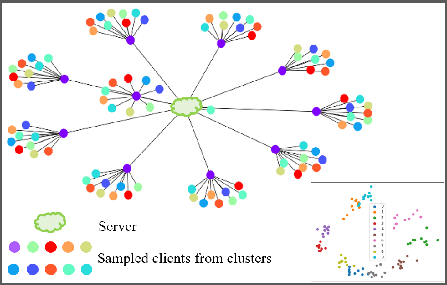}%
\label{subfig:topology555}}
\hfil
\subfloat[CIFAR10 ($C=10$)]{\includegraphics[width=1.8in]{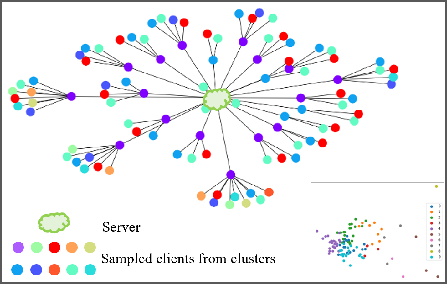}%
\label{subfig:topology666}}

\subfloat[CINIC10 ($C=2$)]{\includegraphics[width=1.8in]{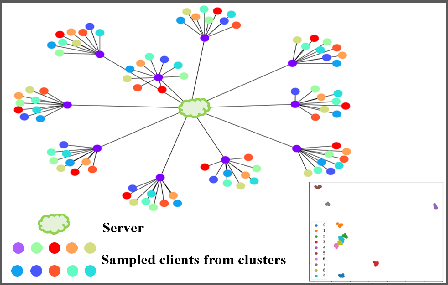}%
\label{subfig:topology777}}
\hfil
\subfloat[CINIC10 ($C=2$)]{\includegraphics[width=1.8in]{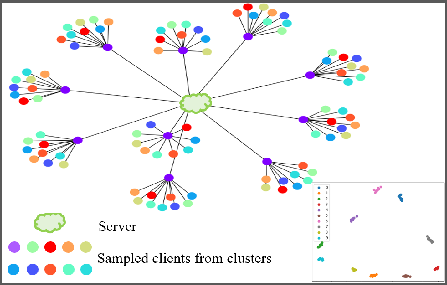}%
\label{subfig:topology888}}
\hfil
\subfloat[CINIC10 ($C=10$)]{\includegraphics[width=1.8in]{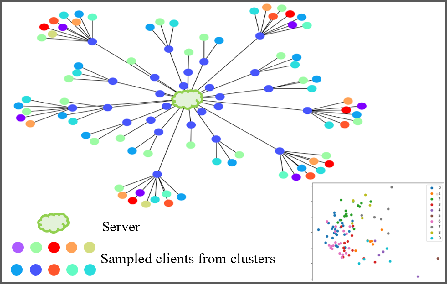}%
\label{subfig:topology999}}

\caption{Heterogeneous topology formed by client clustering, where $N=100$, $K=10$. The bottom right corner of each subfigure shows the results of homogeneous clustering, where different colors represent clients from different homogeneous clusters. In each subfigure, the clients connected with the server are the clusters heads. Each cluster head and its connected clients construct a heterogeneous cluster.}
\label{topologykk}
\end{figure*}

\subsubsection{Baselines}\label{sec:baselines}
To evaluate the performance of our proposed HFLDD, we select five typical baseline methods for comparison:
\begin{itemize}
    \item FedAvg \cite{mcmahan2017communication}, a classic FL framework, where each client conducts local training and the server aggregates all the local models for global model update;
    \textcolor{black}{
    \item FedProx \cite{li2020federated}, a classic regularization-base FL framework, which adopt regularization terms to adjust local updates, and mitigate client drift on non-IID data;
    }
    \item FedSeq \cite{chen2023fedseq}, a classic HFL framework, which utilizes client clustering and sequential training to enhance model performance;
    \textcolor{black}{
    \item  DynaFed \cite{pi2023dynafed}, a dataset distillation-based FL framework, which synthesizes distilled data by matching the training trajectory of the global model on the server, and then fine-tunes the aggregated model using the generated data}
   \item Centralized learning (CL), which can provide the performance upper bound for model training.
\end{itemize}

\subsubsection{Implementation Details}
Our experiments are carried out on a machine equipped with a Tesla V100 GPU and implemented using PyTorch. 
We fix the number of clients $N$ to be 100. The parameters involved in the pre-training phase are set as follows: learning rate $\eta=0.01$, steps of local model updates $E_1=10$, a batch size of 64, and the SGD optimizer. 
To construct the heterogeneous topology, we set the clustering of K-Means with $K=10$.
The size of the distilled data, denoted by $|\tilde{\mathcal{D}_i}|$, is set to 400, which means that each client generates 400 distilled data. 
\textcolor{black}{
The size of the distilled dataset for each client, denoted by $|\tilde{\mathcal{D}}_i|$, is set to 400 across all datasets, meaning that each client generates 400 distilled samples during the experimental evaluation.}
The learning rate for optimizing the distilled data is 0.004, and the maximum number of distillation iterations is 3000. In each iteration, a minibatch of 10 samples is drawn from the original dataset to compute the kernel regression loss and update the distilled data. 
\textcolor{black}{
The scale-invariant regularization coefficient for the kernel loss is fixed at $\lambda = 1 \times 10^{-6}$, following the setting in KIP~\cite{nguyen2021dataset}. This value is adopted to ensure numerical stability during kernel matrix inversion, while exerting negligible influence on the regression solution.
}
Following the transmission of the distilled data, the global model training begins. Each cluster head conducts local training with $E_2=2$ and a batch size of 32. The server performs global model updates for $T=300$ rounds. All other configurations are consistent with the pretraining phase.

To ensure a fair comparison, FedAvg and FedSeq are configured with the same number of global update rounds and local update epochs as HFLDD. Furthermore, in FedSeq, the number of clusters is set to 10, with each cluster containing 10 clients.
For DynaFed, the server synthesizes 15 distilled data per class. The global model is then fine-tuned for 20 epochs using a batch size of 256. For centralized learning, the training batch size is 64, and the learning rate is 0.01.

\subsection{Training Performance}\label{subsec:robust-training} 
To comprehensively show the training performance of the proposed HFLDD, we compare HFLDD with the baseline methods in both non-IID and IID scenarios.

\subsubsection{Non-IID Scenario} We consider two severe non-IID settings: 1) $C=1$, where each client holds only one class of data, making it the most challenging setting; 2) $C=2$, where each client holds two classes of data. The data labels among different clients are still severely unbalanced, but partially overlap.  

In Fig. \ref{fig:accuracy}, we compare the training performance of HFLDD with the baseline methods on the datasets of MNIST, CIFAR10, and CINIC10, respectively. When $C=1$, as shown in Fig. \ref{subfig:accuracy1}, Fig. \ref{subfig:accuracy4}, and Fig. \ref{subfig:accuracy7}, it can be observed that our proposed HFLDD outperforms FedAvg, FedSeq, and DynaFed on all datasets.
Specifically, 
on CIFAR10, HFLDD achieves an accuracy of 74.0\% after 300 communication rounds, significantly outperforming FedAvg at 27.0\%, FedProx at 26.3\%, FedSeq at 19.6\%, and DynaFed at 55.7\%.
On CINIC10, HFLDD achieves an accuracy of 56.7\%. Compared to FedAvg, FedProx, FedSeq, and DynaFed, it shows relative improvements of 39.3\%, 57.1\%, 8.8\%, and 41.4\%, respectively.
On MNIST, although the final accuracy of HFLDD is similar to that of FedSeq, it is clear that HFLDD converges faster. DynaFed also achieves a similar final accuracy, but the use of distilled data introduces instability in the later stages of training.
When $C=2$,  as shown in Fig. \ref{subfig:accuracy5} and Fig. \ref{subfig:accuracy8}, HFLDD achieves final test accuracies of 78.5\% on CIFAR10 and 59.5\% on CINIC10, respectively, both outperform FedAvg, FedProx, FedSeq and DynaFed. 
On MNIST, as shown in Fig. \ref{subfig:accuracy2}, HFLDD, FedSeq, and DynaFed achieve nearly identical training results, and all outperform FedAvg and FedProx. 
\textcolor{black}{
This phenomenon is mainly attributed to the simplicity of the MNIST dataset, which enables different methods to easily learn discriminative features.}
Additionally, under the non-IID settings described above, the performance gap between HFLDD and CL is the smallest.

Figs. \ref{subfig:topology111}, \ref{subfig:topology222}, \ref{subfig:topology444}, \ref{subfig:topology555}, \ref{subfig:topology777} and \ref{subfig:topology888} specifically illustrate the heterogeneous topology formed by HFLDD in the non-IID scenario. 
In Fig. \ref{subfig:topology444},  Fig. \ref{subfig:topology777} and Fig. \ref{subfig:topology888}, the numbers of clients in all heterogeneous clusters are equal. 
In Figs. \ref{subfig:topology111}, \ref{subfig:topology222}, and \ref{subfig:topology555}, a few isolated nodes appear in the heterogeneous topology. However, they do not significantly affect overall performance. This is because the data distribution in most heterogeneous clusters is approximately the same as the global data distribution, effectively compensating for overall performance.


\begin{figure*}[!t]
\centering
\subfloat[MNIST]{\includegraphics[width=0.3\linewidth]{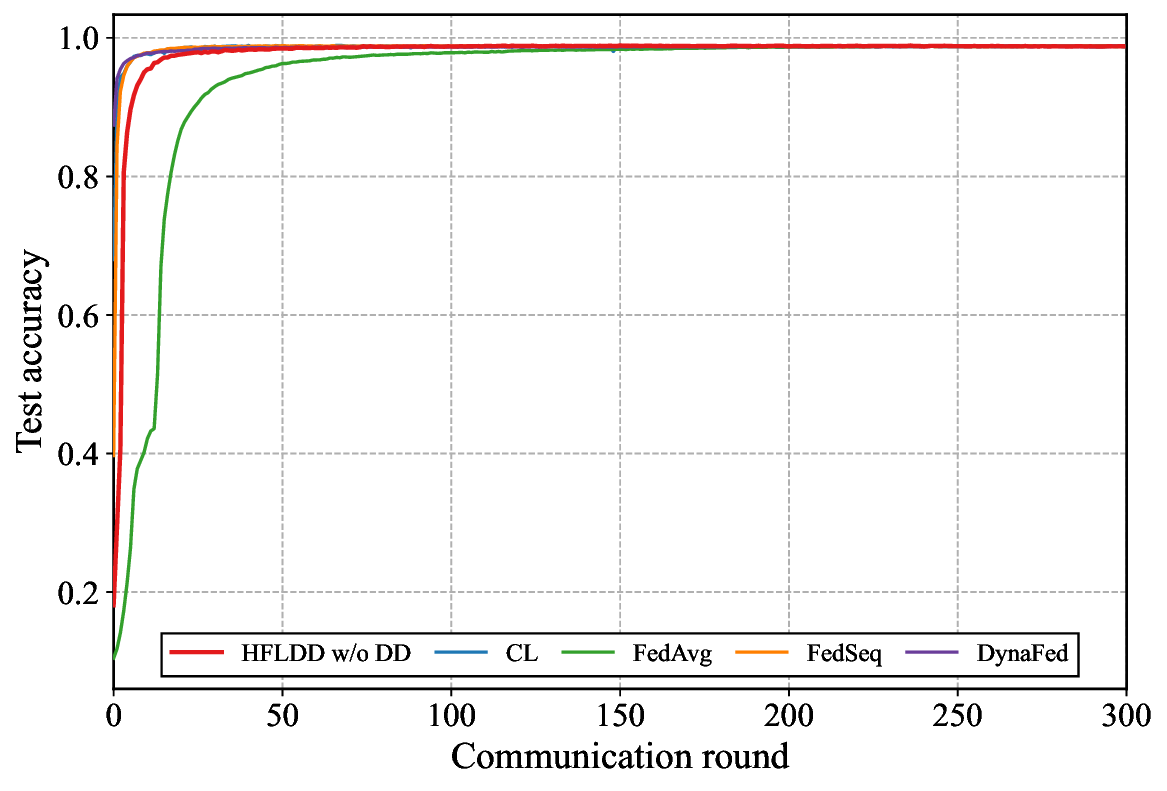}\label{subfig:accuracykk1}}%
\hfil
\subfloat[CIFAR10]{\includegraphics[width=0.3\linewidth]{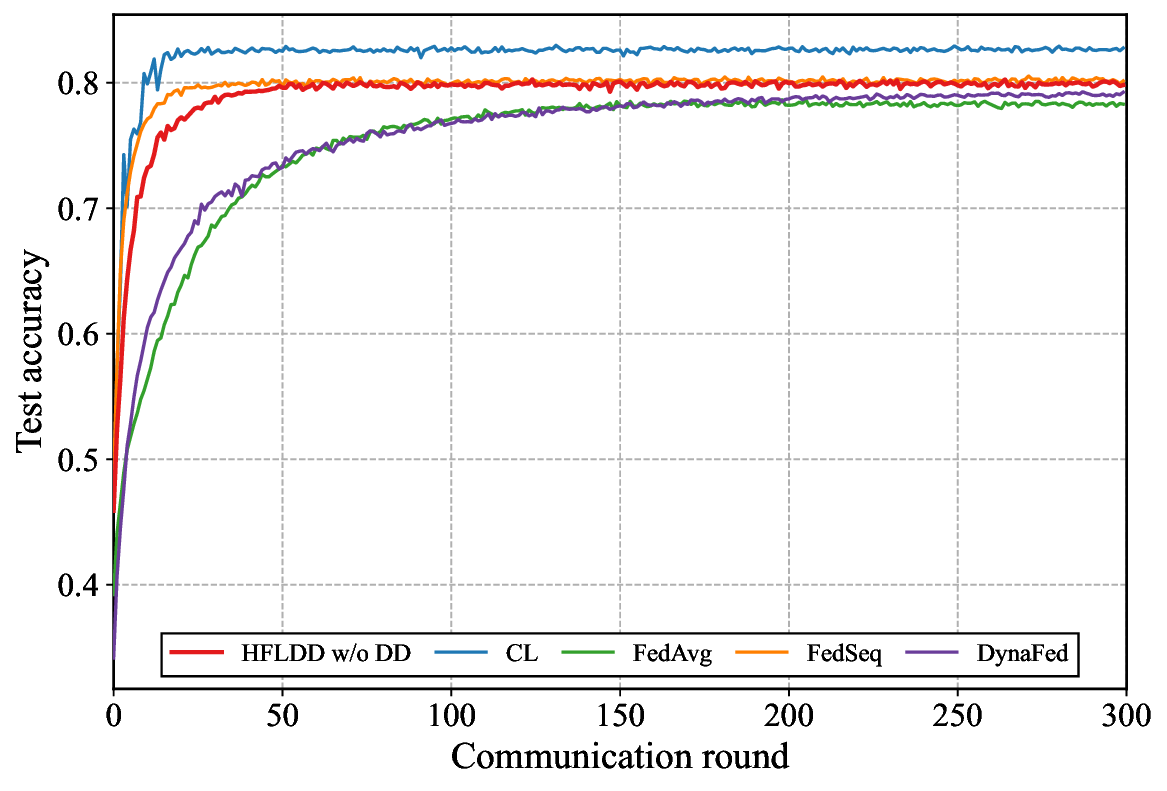}\label{subfig:accuracykk2}}%
\hfil
\subfloat[CINIC10]{\includegraphics[width=0.3\linewidth]{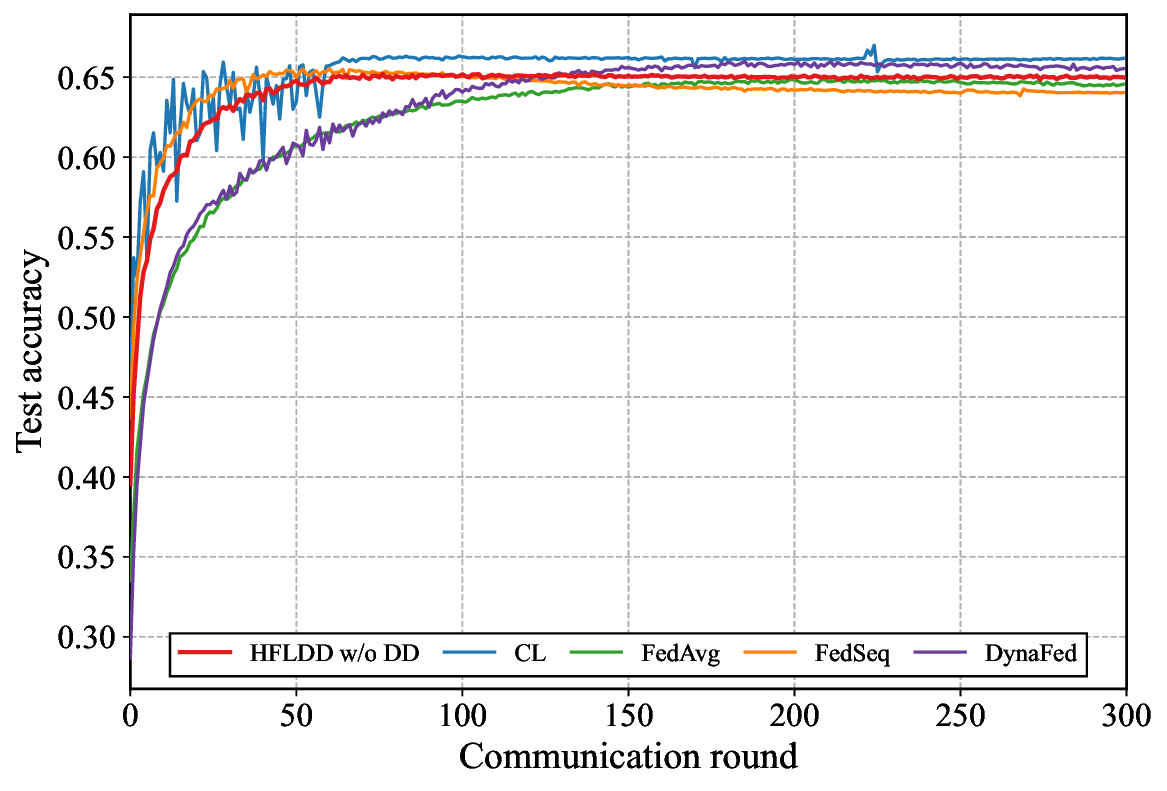}\label{subfig:accuracykk3}}%

\caption{\textcolor{black}{Performance comparison of HFLDD without dataset distillation (HFLDD w/o DD) with baseline methods on three datasets when $C=10$.}}
\label{fig:clusterinfluence}
\end{figure*}

\subsubsection{IID Scenario} In this case, we set $C=10$, where each client holds all ten data classes, and the data labels among all clients are balanced, which significantly reduces the benefit of cross-client knowledge aggregation.
\textcolor{black}{
On MNIST, as shown in Fig.~\ref{subfig:accuracy3}, all methods achieve nearly identical test accuracies, which can also be attributed to the simplicity of data.}
On CIFAR10 and CINIC10, as shown in Fig. \ref{subfig:accuracy6} and Fig. \ref{subfig:accuracy9}, HFLDD exhibits slightly inferior performance compared to some baseline methods. However, it remains competitive in the IID setting.

This performance gap can be attributed to two main factors. First, HFLDD trains the global model using
a combination of the local data from the cluster head and the distilled data collected from its cluster members. 
\textcolor{black}{
The dataset distillation process inevitably introduces information loss. This effect becomes more noticeable in the IID scenario, where baseline methods can already  exploit balanced local data.
}
It is important to emphasize that this information loss is not unique to the IID case but exists across all levels of data heterogeneity. In non-IID settings ($C=1$ or $C=2$), HFLDD still outperforms the baseline methods even with relatively small distilled data. The impact of distilled data size will be shown in Section~\ref{Impact of distilled data size}.

Second, under IID condition where each client holds data from all 10 classes, 
as shown in Figs.~\ref{subfig:topology333}, \ref{subfig:topology666} and \ref{subfig:topology999}, the number of clients in each heterogeneous cluster is imbalanced.
After cluster heads collect the distilled data, each of them clearly holds all 10 classes of data, meaning that the label distributions across cluster heads are identical. However, due to the different number of clients in each cluster, the amount of data held by each cluster head differs, which leads to quantity skew. As discussed in previous work \cite{li2022federated}, such an imbalance has a minor effect on performance when weighted averaging is used during aggregation.
\textcolor{black}{
To further assess its influence in our framework and to isolate the impact of dataset distillation, }
we conduct an ablation experiment where the distilled data transmission step was replaced by direct transmission of the original local data to the cluster head. This variant, denoted by HFLDD w/o DD, removes the influence of dataset distillation. 
As shown in Fig. \ref{fig:clusterinfluence}, HFLDD w/o DD achieves a performance comparable to or even better than the baseline methods in all three datasets, demonstrating that the impact of the cluster size imbalance caused by clustering is indeed minor.

\textcolor{black}{
The above analysis and experimental evidence indicate that the observed performance gap between HFLDD and other baseline methods under the IID condition is primarily attributable to the information loss from dataset distillation, rather than the clustering mechanism.}

\begin{table*}[htbp]
\centering
\caption{
\textcolor{black}{Communication cost when achieving given accuracy on MNIST under IID condition}}
\label{mnisttraffic111}
\begin{tabular}{clccccc}
\toprule
\multicolumn{2}{c}{Metrics} & HFLDD & FedAvg & FedProx  & FedSeq & DynaFed\\
\midrule
\multirow{2}{*}{Accuracy=80\%} & rounds/Accuracy & 5/80.63\% & 17/80.54\% & 20/80.6\% & 1/84.32\% & 2/85.1\%\\
                               & Traffic (MB)    & 58.8M    & 593.15M & 694.8M  & 39.0M  & 84.7M  \\
\midrule
\multicolumn{2}{c}{Ratio}     & 1X              & 10.1X   & 11.8X  & 0.7X    & 1.4X  \\
\bottomrule
\end{tabular}
\end{table*}

\begin{table*}[htbp]
\centering
\caption{\textcolor{black}{Communication cost when achieving given accuracy on CIFAR10 under IID conditions}}
\label{cifar10traffic111}
\begin{tabular}{clccccc}
\toprule
\multicolumn{2}{c}{Metrics} & HFLDD & FedAvg & FedProx  & FedSeq & DynaFed\\
\midrule
\multirow{2}{*}{Accuracy=70\%} & rounds/Accuracy & 10/70.5\% & 35/70.3\% & 43/70.3\% & 4/71.2\% & 30/70.4\% \\
                               & Traffic (MB)    & 1973.7M   & 27630.4M  & 33856.9M & 2296.0M & 23738.7M \\
\midrule
\multicolumn{2}{c}{Ratio}     & 1X              &14.0X   &17.2X & 1.2X   &12.0X \\
\bottomrule
\end{tabular}
\end{table*}

\begin{table*}[htbp]
\centering
\caption{\textcolor{black}{Communication cost when achieving given accuracy on CINIC10 under IID conditions}}
\label{cinic10traffic111}
\begin{tabular}{clccccc}
\toprule
\multicolumn{2}{c}{Metrics} & HFLDD & FedAvg & FedProx  & FedSeq & DynaFed \\
\midrule
\multirow{2}{*}{Accuracy=60\%} & rounds/Accuracy & 40/60.1\% & 44/60.0\% & 117/60.1\% & 11/60.7\% & 50/60.2\% \\
                               & Traffic (MB)    & 2661.4M   & 10864.6M & 28687.4M & 1745.7M  & 12329.5M\\
\midrule
\multicolumn{2}{c}{Ratio}     & 1X              & 4.1X  & 10.8X  & 0.7X  & 4.6X  \\
\bottomrule
\end{tabular}
\end{table*}

\begin{figure*}[!htbp]
\centering
\centering
\subfloat[MNIST]{\includegraphics[width=0.3\linewidth]{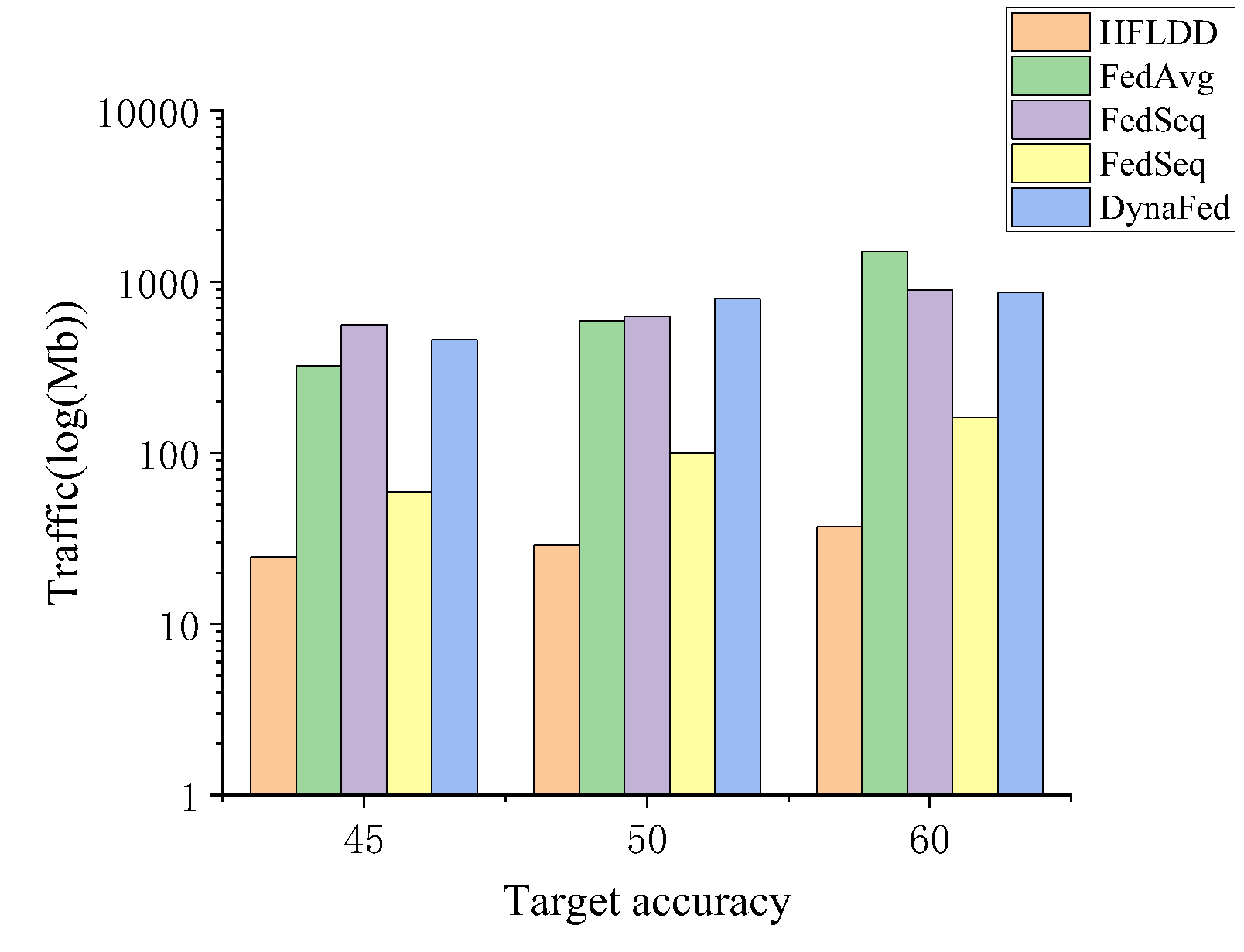}%
\label{subfig:TRAFFICaccuracy1}}
\hfil
\subfloat[CIFAR10]{\includegraphics[width=0.3\linewidth]{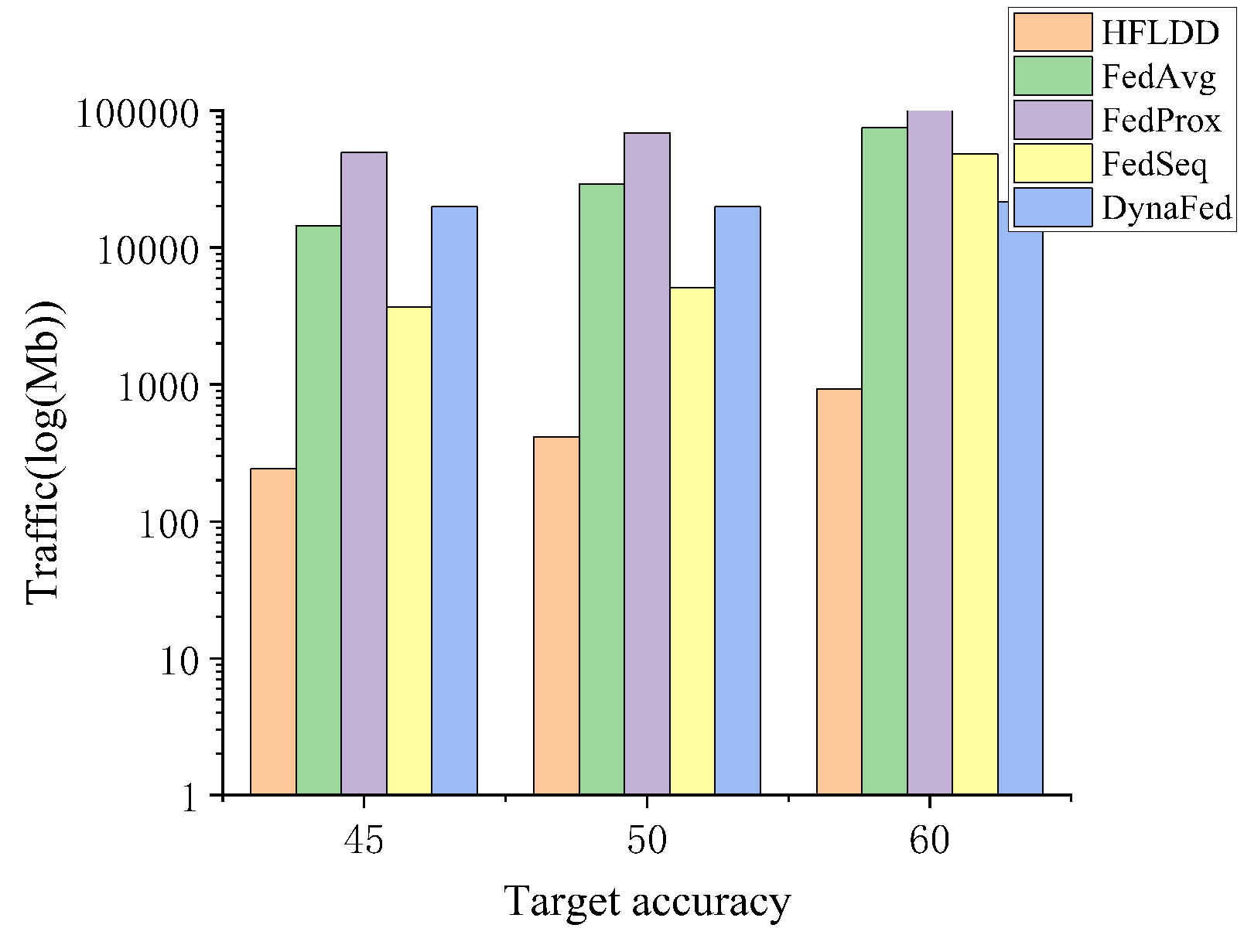}%
\label{subfig:TRAFFICaccuracy2}}
\hfil
\subfloat[CINIC10]{\includegraphics[width=0.3\linewidth]{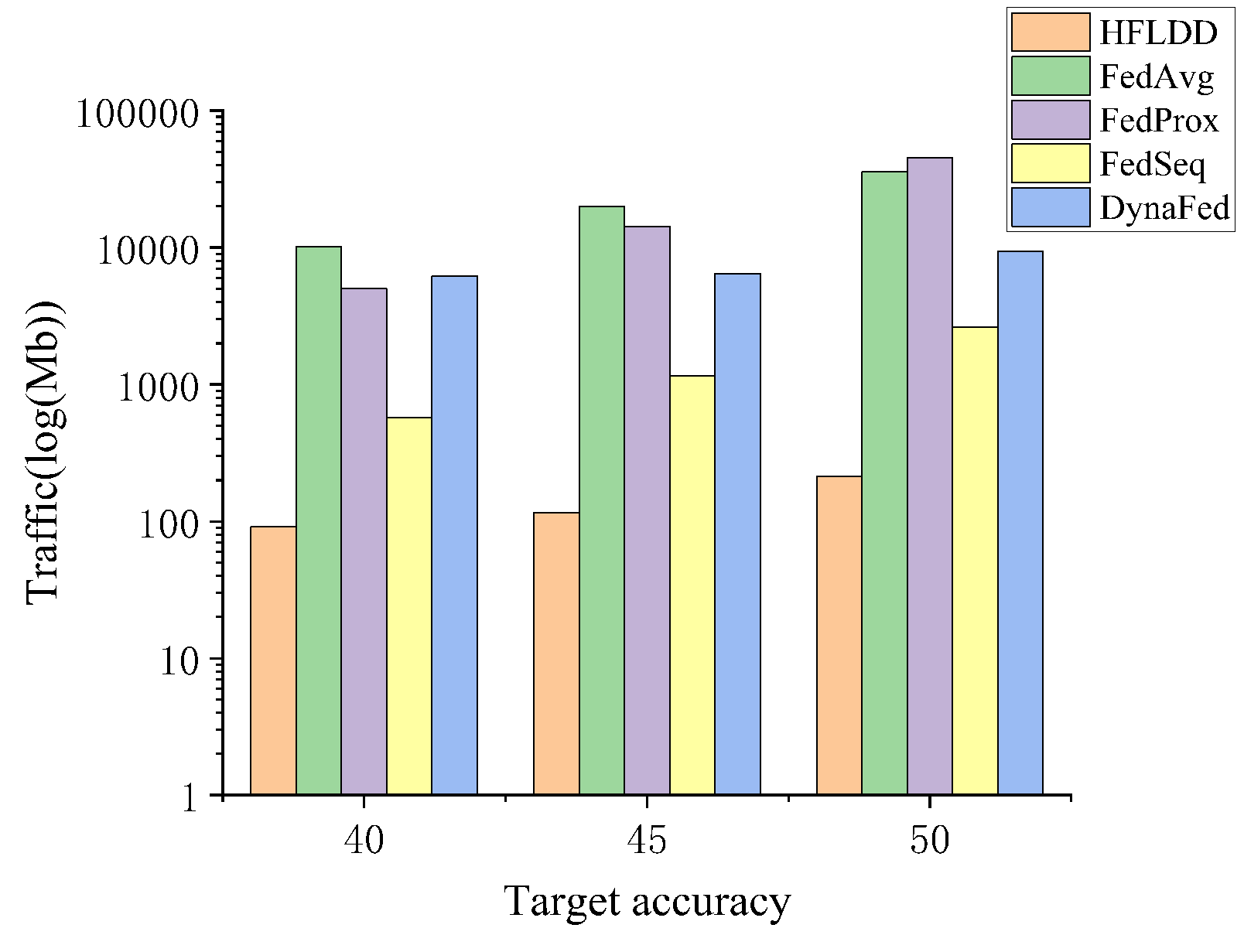}%
\label{subfig:TRAFFICaccuracy3}}
\caption{\textcolor{black}{Communication traffic consumption of three methods when achieving given target accuracy under non-IID conditions $(C=2)$.}
}
\label{trafficlabel2}
\end{figure*}

\subsection{Communication Cost} \label{Communication Efficiency}

\textcolor{black}{
The analytical communication cost of HFLDD is detailed in Eq.~\eqref{eq:communicationhfldd}, which accounts for soft labels uploading, distilled data transmission within each cluster, and communication between the server and cluster heads.
}
In this part, we compare the communication cost of HFLDD with that of baseline methods, including FedAvg, FedProx, FedSeq, and DynaFed.

The communication costs of FedProx and DynaFed over $T$ rounds are the same as those of FedAvg, which is given by Eq. \eqref{eq3}. For FedSeq, the communication cost comes from parameter sharing between clients within cluster and across clusters, it can be expressed as 
\begin{equation}
\mathcal{C}_\textbf{FedSeq}\,\,=\,\,O\cdot |\omega |\cdot B_1\cdot \left( \left( 2T-1 \right) +T\cdot J \right),
\end{equation}
where $O$ is the number of clusters and $J$ is the number of clients in each cluster.

Considering that when training converges, the communication rounds and test accuracies of different methods are different, it would be unfair to compare the communication cost under a given communication rounds. Therefore, we compare the number of communication rounds and the total communication cost required by different methods to achieve a certain test accuracy.


In the IID scenario, the target test accuracies for MNIST, CIFAR10, and CINIC10 are set to 80\%, 70\% and 60\%, respectively. Table~\ref{mnisttraffic111} to Table~\ref{cinic10traffic111} present the number of communication rounds and the total communication costs required by different methods to achieve these target accuracies. The ratio indicates the relative communication cost of each method compared to the HFLDD. Experimental results show that HFLDD incurs slightly higher communication cost than FedSeq on MNIST and CINIC10, but remains lower than that of FedAvg, FedProx, and DynaFed. On CIFAR10, the communication cost of HFLDD is the lowest among all the methods.


In the Non-IID scenario, when $C=1$, due to the severe imbalance of data labels, the baseline methods may not converge and it is infeasible to find a proper target test accuracy that adequately reflects the communication cost of each method. 
Therefore, we consider the case when $C=2$. Fig. \ref{trafficlabel2} compares the communication costs of different methods to achieve given target accuracies on MNIST, CIFAR10, and CINIC10. Since the communication cost of HFLDD is significantly lower than that of the baseline methods in this scenario, we use a logarithmic scale (base 10) on the y-axis. From the figure, it can be observed 
that as the target accuracy increases, HFLDD consistently maintains the lowest communication cost.

\begin{figure*}[!htbp]
\centering
\subfloat[$C=1$]{\includegraphics[width=0.3\linewidth]{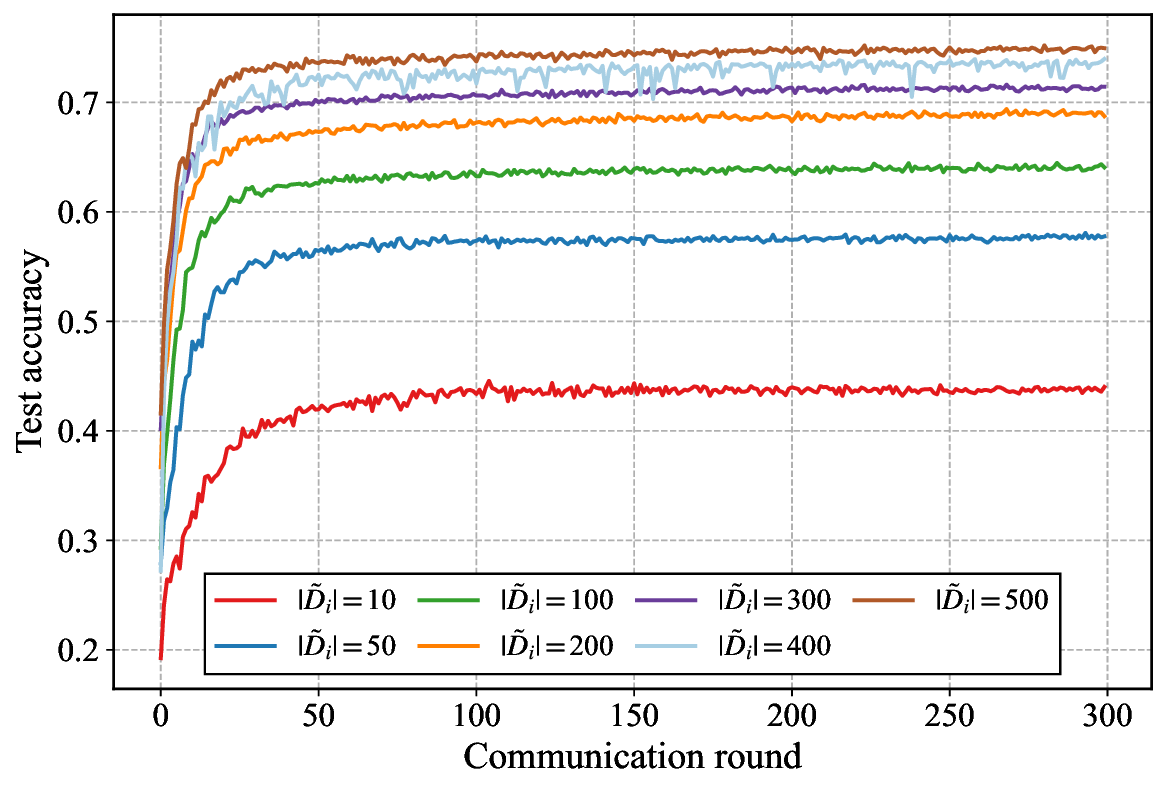}%
\label{subfig:ipcaccuracy1}}
\hfil
\subfloat[$C=2$]{\includegraphics[width=0.3\linewidth]{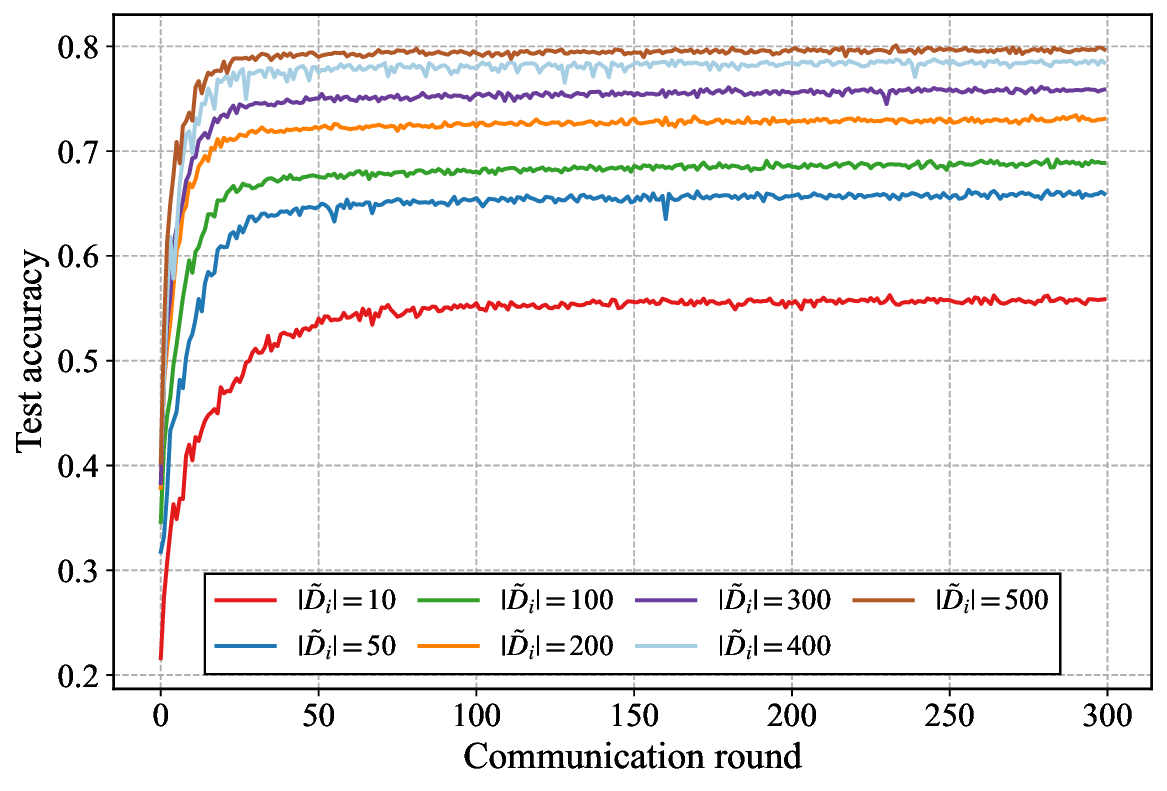}%
\label{subfig:ipcaccuracy2}}
\hfil
\subfloat[$C=10$]{\includegraphics[width=0.3\linewidth]{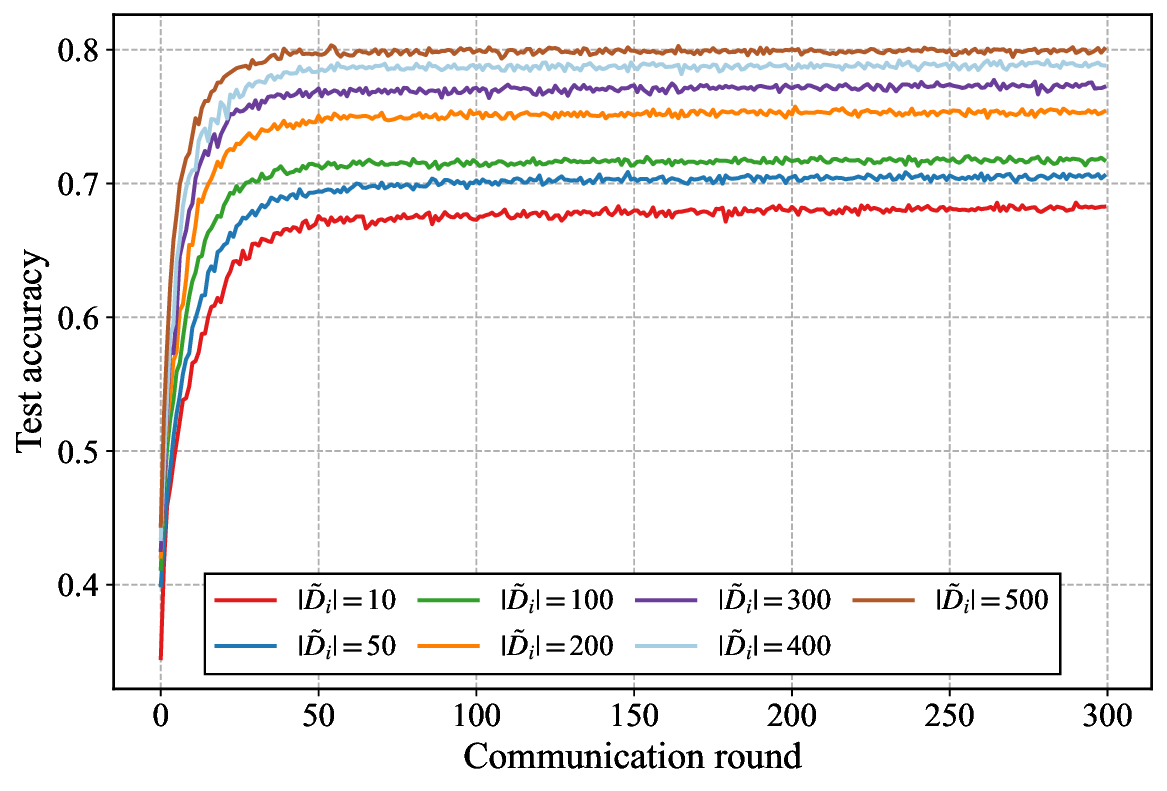}%
\label{subfig:ipcaccuracy3}}

\caption{\textcolor{black}{
Learning curves of the global model with different distilled data sizes across varying
data heterogeneity on CIFAR10.
}
}
\label{fig:ipcnumber}
\end{figure*}


\textcolor{black}{
\subsection{Impact of distilled data size} \label{Impact of distilled data size}
The size of the distilled data significantly affects the amount of information retained from the local data of cluster members, which directly influences the performance of the global model. To investigate this impact, we evaluate the convergence behavior of HFLDD under varying levels of data heterogeneity on CIFAR10, using different distilled data sizes.
As shown in Fig.~\ref{fig:ipcnumber}, increasing the distilled data size leads to notable improvements in both convergence speed and final test accuracy. This observation aligns with the theoretical expectation: a larger distilled data can better approximate the original data distribution, thereby reducing the distillation approximation error and improving the performance of the global model.
}

\begin{table}[!b]
\centering
\caption{Test accuracies with different model architectures on CIFAR10}
\begin{tabular}{llccc}
\toprule
\textbf{Method} & \textbf{$C$} & \multicolumn{3}{c}{\textbf{Model}} \\
\cline{3-5}
 & & \textbf{MobileNetV2} & \textbf{ResNet18} & \textbf{ShuffleNet} \\
\midrule
\multirow{3}{*}{HFLDD} 
& $C=1$  & 65.8\% & 71.0\% & 58.3\% \\
& $C=2$  & 71.2\% & 73.5\% & 66.1\% \\
& $C=10$ & 71.5\% & 74.6\% & 71.9\% \\
\midrule
\multirow{3}{*}{FedSeq} 
& $C=1$  & 11.5\% & 9.8\% & 11.4\% \\
& $C=2$  & 30.1\% & 27.9\% & 26.8\% \\
& $C=10$ & 72.9\% & 74.9\% & 73.6\% \\
\midrule
\multirow{3}{*}{FedAvg} 
& $C=1$  & 11.0\% & 11.6\% & 10.3\% \\
& $C=2$  & 30.2\% & 32.6\% & 45.6\% \\
& $C=10$ & 68.9\% & 61.0\% & 68.5\% \\
\midrule
\multirow{3}{*}{DynaFed} 
& $C=1$  & 44.8\% & 55.6\% & 32.7\% \\
& $C=2$  & 57.9\% & 73.7\% & 45.7\% \\
& $C=10$ & 62.9\% & 78.6\% & 68.6\% \\
\bottomrule
\end{tabular}
\label{tab:method_model_comparison}
\end{table}

\textcolor{black}{
\subsection{Impact of Network Architecture} \label{Impact of different model}
To evaluate the generalizability and adaptability of our proposed framework across various network architectures, we conducted additional experiments using a diverse set of models on CIFAR10, including MobileNetV2\cite{sandler2018mobilenetv2}, ResNet18\cite{he2016deep}, and ShuffleNet\cite{zhang2018shufflenet}. These models cover a broad spectrum of architectural characteristics, such as lightweight mobile models and deeper residual networks.
}

\textcolor{black}{
Table~\ref{tab:method_model_comparison} lists the test accuracies of different methods using the above three models under different levels of client heterogeneity. Overall, HFLDD exhibits stable and competitive performance across all model architectures and heterogeneity settings. Under severe non-IID settings, i.e., when $C=1$ and $C=2$, the test accuracies of HFLDD are largely higher than FedAvg and FedSeq, and mostly higher than that of DynaFed. Under the IID setting, i.e., when $C=10$, HFLDD remains competitive with the baseline methods, demonstrating strong adaptability across different model architectures. 
}

\begin{table*}[t]
\centering
\caption{
\textcolor{black}{
The impact of $K$ on the test accuracy of HFLDD
}
}
\label{IMPACTk}
\begin{tabular}{
  >{\centering\arraybackslash}m{2.5cm}  
  >{\centering\arraybackslash}m{1.5cm}
  >{\centering\arraybackslash}m{1.5cm}
  >{\centering\arraybackslash}m{1.5cm}
  >{\centering\arraybackslash}m{1.5cm}
  >{\centering\arraybackslash}m{1.5cm}
  >{\centering\arraybackslash}m{1.5cm}
  >{\centering\arraybackslash}m{1.5cm}
}
\toprule
{The Value of K} & {5} & {6} & {7} & {8} & {9} & {10} & {11} \\
\midrule
MNIST ($C=1$)  & 95.2\% & 97.1\% & 97.3\% & 98.0\% & 98.0\% & 98.1\% & 98.1\% \\
MNIST ($C=2$)  & 98.3\% & 98.0\% & 98.3\% & 98.3\% & 98.1\% & 98.3\% & 98.1\% \\
MNIST ($C=10$) & 98.5\% & 98.3\% & 98.5\% & 98.4\% & 98.3\% & 98.4\% & 98.4\% \\
CIFAR10 ($C=1$)  & 64.1\% & 71.6\% & 70.3\% & 68.4\% & 71.5\% & 74.0\% & 71.0\% \\
CIFAR10 ($C=2$)  & 79.0\% & 78.1\% & 79.0\% & 77.5\% & 77.3\% & 78.5\% & 76.6\% \\
CIFAR10 ($C=10$) & 78.5\% & 79.4\% & 79.3\% & 79.7\% & 79.1\% & 78.8\% & 77.9\% \\
CINIC10 ($C=1$)  & 51.8\% & 51.5\% & 55.3\% & 54.8\% & 54.4\% & 56.5\% & 55.2\% \\
CINIC10 ($C=2$)  & 57.9\% & 58.1\% & 58.5\% & 57.8\% & 57.5\% & 59.5\% & 59.7\% \\
CINIC10 ($C=10$) & 62.8\% & 63.5\% & 63.7\% & 62.8\% & 62.9\% & 62.6\% & 62.0\% \\
\bottomrule
\end{tabular}
\end{table*}



\subsection{\texorpdfstring{Impact of $K$ in K-Means clustering}{Impact of K in K-Means clustering}} \label{Impact of different K}
In this part, we evaluate the impact of $K$ on the homogeneous clustering of K-Means clients on the performance of HFLDD. Table \ref{IMPACTk} shows the test accuracies of HFLDD in MNIST, CIFAR10, and CINIC10 with different values of $K$. When $C=1$, on CIFAR10 and CINIC10, HFLDD achieves the highest test accuracy when $K=10$, and shows a noticeable decrease
as $K$ decreases or increases. However, in MNIST, the test accuracy
remains relatively unchanged as $K$ varies. This is because MNIST is a relatively simple dataset, its data patterns and features are easy to capture and understand. In this scenario, even if an inappropriate value of $K$ is selected, HFLDD can still easily learn the features in the MNIST dataset.

When $C=2$ and $C=10$, there is overlap in the label classes between different clients, leading to a certain degree of overlap in the feature space. The test accuracies on the three datasets remain relatively stable as the value of $K$ changes. We can see that when data heterogeneity is not particularly severe, HFLDD exhibits robustness to the value of $K$.

\begin{figure}[!htbp]
\centering
\subfloat[]{\includegraphics[width=0.45\linewidth]{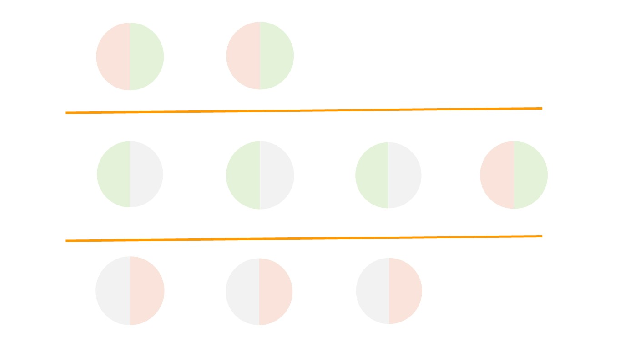}%
\label{mnisttraffic99}}
\hfil
\subfloat[]{\includegraphics[width=0.55\linewidth]{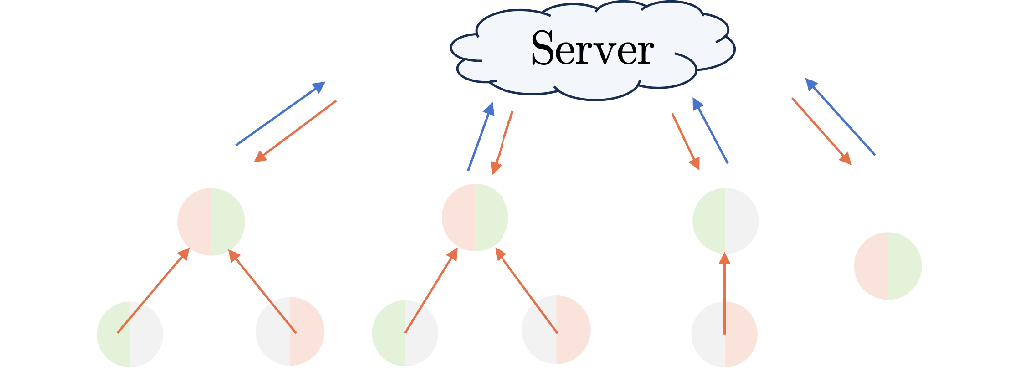}%
\label{cifar10traffic99}}
\caption{(a) shows homogeneous clusters, where nine clients are grouped into three 
homogeneous clusters. (b) shows the heterogeneous topology diagram, where the heterogeneous clusters with only one or two clients cannot construct datasets with data distributions similar to the global distribution.} 
\label{fig:topology3}
\end{figure}

\section{Discussion}\label{sec:discuss}
\textcolor{black}{
It is important to note that the actual scenario differs from the example provided above, 
the data heterogeneity among clients' data types is highly complex. After sampling and reorganizing homogeneous clusters, it is highly possible to have a limited number of clients in some heterogeneous clusters. There may even be many isolated nodes considered as a single heterogeneous cluster. This leads to an uneven number of clients among clusters. In decentralized settings, if these uneven clusters are linked, as in Ref. \cite{abebe2023optimizing}, to form a ring-shaped topology for collaborative training, it will lead to a significant decrease in convergence speed and test accuracy. The most intuitive approach to address this issue is to directly remove heterogeneous clusters with isolated or a very small number of clients from training. However, this often results in worse training performance, which is unable to fully utilize the data in each client.
}

\textcolor{black}{
For better understanding of this issue in HFL, we let each client hold two different classes of data. Due to the overlap of data classes between clients, it is likely to form homogeneous clusters as illustrated in Fig. \ref{fig:topology3}a. In this case, through sampling and reorganization, the constructed heterogeneous clusters have unbalanced client numbers, as shown in Fig. \ref{fig:topology3}b. After transmitting the distilled data, heterogeneous clusters with fewer clients cannot obtain a dataset whose distribution approximates the global distribution.}




\textcolor{black}{
Although some heterogeneous clusters may have uneven client sizes and lack complete label coverage, HFLDD is designed to robustly tolerate such imperfections. This robustness is mainly attributed to three factors.
}
\textcolor{black}{
First, it promotes statistical diversity by sampling clients from different homogeneous groups. Although full label coverage within each heterogeneous cluster is not guaranteed, this strategy encourages the inclusion of diverse label types. As a result, many clusters exhibit approximately balanced label distributions in practice, which is both realistic and sufficient for achieving robust training, particularly under severe non-IID conditions.
Second, the presence of a central server makes data exchange and collaboration more flexible. Each client participates in at least one heterogeneous cluster, and through multiround global aggregation, the distilled knowledge from all clients is eventually integrated into the global model. This ensures that no client is excluded from the learning process.
Third, while uneven cluster sizes result in different numbers of distilled data across cluster heads, our experiments show that such an imbalance has a negligible effect on the final model performance. 
Therefore, HFLDD maintains stable performance despite inevitable imperfections in cluster formation, making it particularly suitable for real-world implementation.
}


\section{Conclusion}\label{sec:conclusion}
In this paper, we propose a communication-efficient HFL framework that addresses the challenge of distributed learning with non-IID data by leveraging dataset distillation. Our approach constructs heterogeneous clusters based on the knowledge of the labels of all clients. Within each heterogeneous cluster, members transmit distilled data to a cluster head. This process generates approximately IID datasets among clusters. Extensive experiments demonstrate that our proposed HFLDD outperforms the considered baseline methods and can significantly reduce the impact of non-IID data on training performance with a much smaller communication cost.

 \textcolor{black}{
While the proposed HFLDD framework demonstrates significant advantages in handling non-IID data, it has several limitations that need further investigation. First, HFLDD relies on dataset distillation to compress local data into distilled data, which inevitably introduces information loss, particularly under highly imbalanced or long-tailed label distributions. Second, although distilled data are generally considered to offer privacy protection, potential risks may still exist in highly sensitive applications. Third, the current clustering mechanism does not consider system heterogeneity and communication instability, which may affect cluster formation and overall training performance.
}

 \textcolor{black}{
 We plan to address these limitations in future work by developing lightweight dataset distillation algorithms that are better suited for long-tailed data distributions to reduce information loss and computational overhead, investigating privacy-enhancing techniques for distilled data, and extending our framework by incorporating adaptive clustering strategies and more resilient communication mechanisms that can support dynamic participation and improve resilience to communication instability.
}




\bibliographystyle{IEEEtran}
\bibliography{IEEEabrv.bib, ref.bib}

\end{document}